%% file: pr_main.tex
\documentclass[preprint,review,12pt]{elsarticle}
\biboptions{sort&compress}

\usepackage[T1]{fontenc}
\usepackage{graphicx}
\usepackage{amsfonts}
\usepackage{nicefrac}
\usepackage{microtype}
\usepackage{booktabs}
\usepackage{multirow}
\usepackage{enumitem}
\usepackage{wrapfig}
\usepackage{longtable}
\usepackage{pdflscape}
\usepackage{sidecap}
\usepackage{textcomp}
\usepackage{float}
\usepackage{rotating}
\usepackage{url}
\usepackage[hidelinks]{hyperref}

\input{math_commands.tex}

\usepackage[capitalize]{cleveref}

\providecommand{\maketitlesupplementary}{%
  \clearpage
  \begin{center}\Large\bfseries Appendix\end{center}
  \vspace{1em}}

\journal{Pattern Recognition}

\begin{document}
\let\WriteBookmarks\relax
\def\floatpagepagefraction{1}
\def\textpagefraction{.001}

\begin{frontmatter}

\title{\texorpdfstring{Foveation-Guided Dynamic Token Selection\\ for Robust and Efficient Vision Transformers}{Foveation-Guided Dynamic Token Selection for Robust and Efficient Vision Transformers}}

\author[inst1]{Ibrahim Batuhan Akkaya\corref{cor1}}
\ead{bthakkaya@gmail.com}

\author[inst2]{Kishaan Jeeveswaran}

\author[inst2]{Bahram Zonooz\fnref{equal}}

\author[inst2]{Elahe Arani\fnref{equal}}

\cortext[cor1]{Corresponding author}
\fntext[equal]{Equal contribution}

\affiliation[inst1]{organization={Advanced Research Lab, NavInfo Europe},
            city={Eindhoven},
            postcode={5657 DB},
            country={Netherlands}}

\affiliation[inst2]{organization={Department of Mathematics and Computer Science, Eindhoven University of Technology},
            city={Eindhoven},
            postcode={5612 AZ},
            country={Netherlands}}

\input{sec/0}

\begin{highlights}
\item FDT integrates foveation and fixation into vision transformers.
\item Dynamic fixation selects informative tokens in a single feedforward pass.
\item Foveated tokens encode multi-scale context for adaptive attention.
\item FDT improves robustness without adversarial or corruption training.
\item At 50\% fixation budget, FDT reduces MACs by 34.57\%.
\end{highlights}


\end{frontmatter}

\input{sec/1}
\input{sec/2}
\input{sec/3}
\input{sec/4}
\input{sec/6}
\input{sec/5}


\bibliographystyle{elsarticle-num-names}
\bibliography{tmlr_main}

\appendix
\input{sec/X_supp}

\end{document}

%% file: math_commands.tex
\usepackage{amsmath,amsfonts,bm}

\def\eqref#1{equation~\ref{#1}}

\def\1{\bm{1}}

\DeclareMathAlphabet{\mathsfit}{\encodingdefault}{\sfdefault}{m}{sl}
\SetMathAlphabet{\mathsfit}{bold}{\encodingdefault}{\sfdefault}{bx}{n}

\newcommand{\R}{\mathbb{R}}



%% file: sec/0.tex
\begin{abstract}
The human visual system (HVS) employs foveated sampling and eye movements to achieve efficient perception, conserving both metabolic energy and computational resources.
Drawing inspiration from this robustness and adaptability, we introduce the Foveated Dynamic Transformer (FDT), a foveation-guided dynamic token-selection architecture that integrates these mechanisms into a vision transformer framework.
The FDT exhibits strong resilience to various types of noise and adversarial attacks, despite not being explicitly trained for such challenges. This inherent robustness is achieved through the use of fixation and foveation modules: the fixation module identifies fixation points to filter out irrelevant information, while the foveation module generates foveated embeddings with multi-scale information. At the 50\% fixation-budget setting, FDT achieves higher accuracy than DeiT-S (81.9\% vs. 80.9\%) while reducing multiply-accumulate operations by 34.57\%, highlighting one operating point on its accuracy-efficiency trade-off.
These attributes position FDT as an HVS-inspired step toward artificial neural networks that combine adaptive computation with improved resilience.\footnote{The code will be shared upon acceptance.}
\end{abstract}

%% file: sec/1.tex
\section{Introduction}
\label{sec:intro}

Recent studies indicate that deep neural networks and the human brain interpret the environment differently, with the human visual system (HVS) dynamically filtering task-irrelevant information to focus on potential objects of interest,
a selective mechanism that contributes to perceptual stability and resilience against noisy or misleading inputs
\citep{dodge2017study, JMLR:v20:19-519, DBLP:journals/corr/SzegedyZSBEGF13, carlini2017towards}. The retina contains photoreceptors, with the fovea--a high spatial resolution area--playing a key role in color perception and visual detail recognition \citep{curcio1990human}. The highest photoreceptor density at the fovea decreases with eccentricity, resulting in a variable-resolution image transmitted to the brain, a phenomenon known as foveation, highlighting HVS's multi-resolution perception.
Such mechanisms suggest that vision models could benefit from integrating spatially adaptive processing strategies inspired by the HVS.
In addition to efficiency and robustness, such biologically inspired mechanisms enhance spatial awareness, allowing models to preserve fine details in salient regions while maintaining global scene coherence, similar to how the human visual system balances local and global perception.
Studying HVS to enhance deep neural network design is therefore a promising research avenue for developing intelligent agents.

Approximately $10^7$ to $10^8$ bits of information enter the visual nerve every second in the HVS \citep{itti2001computational}. To manage this data efficiently, the HVS uses saccadic eye movements to direct the fovea to selected targets, creating a detailed scene map from varied resolutions, known as fixation points, and saving computational resources \citep{itti2001computational, bruce2009saliency}. Inspired by the HVS, several studies have incorporated foveation and fixation mechanisms into neural networks \citep{mnih2014recurrent, akbas2017object, thavamani2021fovea}. Existing methods sequentially process the input image, first locating the fixation points and then processing the features around the fixation points. However, these approaches are not optimal for two reasons. First, several inferences are required for each fixation point. Second, they require a fusion mechanism to exploit the collective information acquired from various fixation points.

\begin{figure}
    \centering
\includegraphics[width=\columnwidth]{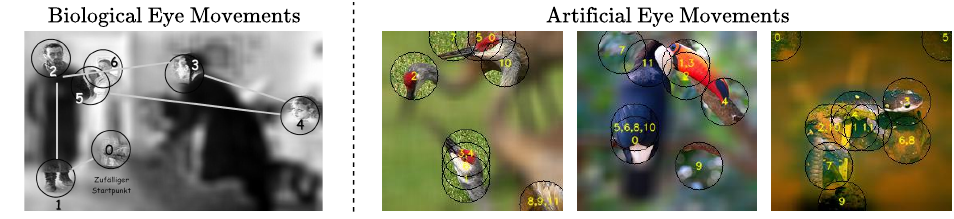}
    \caption{Illustration comparing the fixations of the human visual system (left; \citet{yarbus1967eye}) and an ANN (right; our method). Areas outside the fixations are blurred to highlight regions of interest. The sequence of eye movements is indicated by numbers. Notably, the ANN exhibits overlapping fixation points, as shown by multiple numbers at the same location, separated by commas.}
    \label{fig:biological_and_artificial_fixations}
\end{figure}

We therefore propose a biologically inspired transformer architecture dubbed the Foveated Dynamic Transformer (FDT), comprising foveation and fixation modules that dynamically select multiscale tokens based on the input image. To simulate foveation in HVS, we process input tokens with the foveation module to generate multiscale queries, keys, and values. Inspired by the radial-polar pooling model of foveation proposed in \citep{freeman2011metamers}, foveation module transforms the token into several scales with increasing window size. We employ a dynamic, input-dependent mechanism to simulate eye movement, with the dynamic fixation module producing a fixation map for each token in each transformer block.
Tokens that are not at the fixation point are not processed. The multi-head attention processes only the remaining tokens. The processed tokens that are located at the fixation points are merged with the non-fixated tokens and are sent to the next block. Multiple blocks in transformer process information from multiple fixation points by combining the information transferred from the previous blocks, which enables to implement foveation and fixation mechanisms in single pass.
While inspired by the biological process, our implementation diverges from the iterative fixation control of the HVS, and also from methods like FoveaTer \citep{jonnalagadda2021foveater} that require multiple sequential attention steps to simulate gaze shifts. Instead, we embed both foveation and fixation directly into each transformer block, enabling spatial attention shifts in a single feedforward pass. This approach avoids iterative computation, unlike FoveaTer, and does not rely on token pruning heuristics like DynamicViT \citep{rao2021dynamicvitefficientvisiontransformers}, allowing for end-to-end differentiable and biologically grounded token selection.

Using a DeiT architecture as a baseline, we evaluate our model on an image classification task. We integrate the fixation and foveation modules into the same architecture and evaluate the effectiveness, efficiency, and robustness of our model on the ImageNet100 database. The FDT architecture enhances the robustness of the vision transformers against adversarial attacks, shortcut learning, and natural corruption by 27\%, 6\%, and 3\%, respectively, without being directly trained for these specific challenges. Moreover, FDT achieves a 34\% reduction in computational demand, measured in Multiply-Accumulate operations (MACs), demonstrating its efficiency and effectiveness in processing while maintaining a lean computational footprint.

%% file: sec/2.tex
\section{Related Work}

Several approaches inspired by the functioning of the human visual system have been devised to replicate foveation and eye movement within the domain of computer vision. \citet{mnih2014recurrent} introduced a method that employs a sequence of image movements to aggregate information before classification. This method uses a hard-attention mechanism trained via reinforcement learning to predict fixation points. In a similar vein, \citet{akbas2017object} developed a foveated object detection system that harnesses varying resolutions to comprehensively analyze the entire image, aligning its fovea with regions of interest in the input data. This approach amalgamates data from multiple fixations and leverages peripheral information, similar to the way the human visual system employs contextual cues to guide gaze.


\begin{figure}[t]
    \centering
    \includegraphics[width=\textwidth]{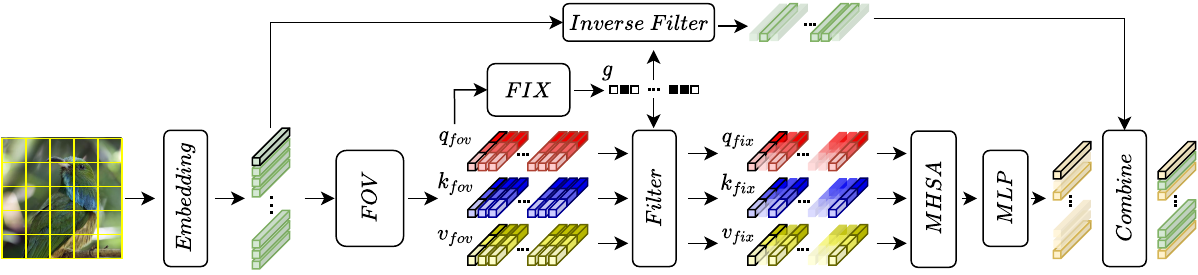}
    \caption{Schematic of the \textit{Foveated Dynamic Transformer (FDT)}, our novel approach inspired by the HVS. FDT integrates fixation and foveation mechanisms within a single processing pass, \textbf{eliminating the need for iterative passes}. Input tokens undergo foveation via the \textit{FOV} module, sampling features at varying resolutions and yielding foveated query, key, and value vectors. The \textit{FIX} module then uses these vectors to identify specific fixation points for targeted token processing. The fixation map $g$ filters foveated features within the \textit{MHSA} module. After processing through an \textit{MLP} block, the architecture combines selectively processed fixated tokens with the remainder, positioning processed tokens at fixation points and unprocessed tokens in their original locations. This HVS-inspired mechanism enhances the efficiency and interpretability of neural networks.}
    \label{fig:main_diagram}
\end{figure}

\citet{lukanov2021biologically} developed an efficient model that incorporates space-variant sampling, mimicking the human retina, and the mechanisms to generate sequences of fixations. They proposed a CNN-based method that uses Foveal Cartesian Geometry (FCG) sampling, as outlined by \citet{martinez2006new}, to compress visual signals. An attention mechanism is employed for ``eye movements" to progressively gather detailed information from a scene. Activation within the feature maps of the final convolutional layer is harnessed to guide the generation of fixation sequences. Additionally, \citet{wang2021use} investigate the role of foveation and saccadic eye movements as biologically inspired proxies for data augmentation in the context of self-supervised learning (SSL). They suggest that foveation through cortical magnification and saccade-like sampling of images can replace conventional SSL augmentations, offering insights into potential biological implementations of self-supervision and challenging spatially uniform processing assumptions in both human and machine vision.

More recently, \citet{jonnalagadda2021foveater} introduced foveation and fixation mechanisms directly into the vision transformer architecture. In this method, the next fixation location is iteratively determined based on the final layer's attention weights. While this approach captured dynamic attention patterns akin to human visual processing, its iterative nature results in inefficiencies during inference. Similarly, \citet{li2025gazeguidedlearningavoidingshortcut} introduced a model that incorporates human gaze sequence using a dual-sequence encoder alongside a vision transformer to correct shortcut biases in visual classification. However, their method required separately modeling gaze data and training an additional gaze encoder module. In contrast, our approach integrates foveation and fixation mechanisms directly into the vision transformer without the need for a separate gaze sequence encoder or auxiliary training components.

Building upon prior biologically inspired studies of foveation and fixation in machine vision \citep{mnih2014recurrent, akbas2017object, deza2020emergent, thavamani2021fovea},
our work extends this line of inquiry to the transformer framework.
A detailed overview of these earlier models is provided in \hyperref[sec:extended_related]{Appendix B}.
In parallel, recent research has focused on improving the computational efficiency of vision transformers through token pruning and slimming strategies
\citep{mao2025efficienttokencompressionvision, yu2023xprunerexplainablepruningvision, rao2021dynamicvitefficientvisiontransformers, kim2022learnedtokenpruningtransformers, wang2021adaptivefocusefficientvideo, tang2022patchslimmingefficientvision, wang2024zerotprunezeroshottokenpruning, zhang2024synergistic, Zheng_2023_ICCV};
we summarize these efficiency-oriented approaches in \hyperref[sec:extended_related]{Appendix B}.

Our approach integrates foveation and fixation directly within each transformer block, enabling multiple spatial attention shifts in a single feed-forward pass without the iterative computation required by methods such as FoveaTer. Unlike approaches that assign continuous importance scores for gradual pruning, our fixation mechanism produces discrete, binary decisions informed by multi-resolution foveated features. This design reflects the all-or-nothing nature of biological fixation, simplifies the fusion of processed and unprocessed tokens, and yields a biologically grounded, computationally efficient alternative to existing token-slimming strategies.


%% file: sec/3.tex
\section{Foveated Dynamic Transformer (FDT)}


\subsection{Foveation Module}

We introduce FDT, an HVS-inspired transformer with two additional modules: Foveation and Fixation. The \textit{Foveation module} embeds input features at multiple resolutions, drawing motivation from multi-resolution processing in the human visual system. The \textit{Fixation module} selectively processes tokens through gaze-inspired sampling, producing binary decisions for each foveated query. The MHSA module then processes a subset of the foveated queries, keys, and values based on the Fixation module's decisions, while the remaining tokens are passed to the next block. The FDT architecture is shown in \cref{fig:main_diagram}.

\begin{figure}
    \centering
    \includegraphics[width=\textwidth]{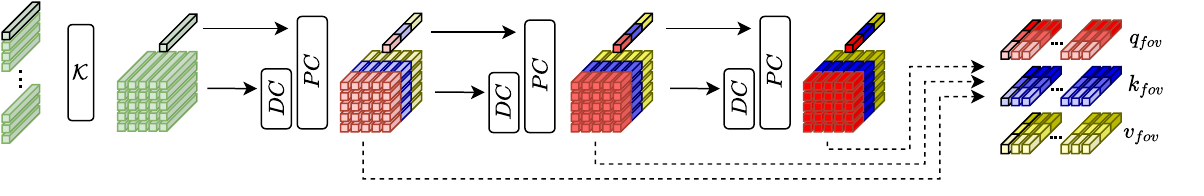}
    \caption{\textbf{Foveation module} comprises three depthwise-separable foveation levels with progressively larger receptive fields. It processes auxiliary and patch tokens using depthwise ($\mathcal{DC}$) separable convolutions and pointwise ($\mathcal{PC}$) convolutions. The module then splits and merges these features to form foveated-query, -key, and -value features, incorporating multi-scale information.}
    \label{fig:foveation}
\end{figure}

The Foveation module in the FDT architecture is designed as a computational analogue of foveated processing by incorporating multi-scale information from neighboring features to generate query, key, and value features (\Cref{fig:foveation}). This mechanism enables each token to aggregate information from progressively larger receptive fields. The Foveation module consists of multiple depthwise separable convolutional layers, each of which progressively extracts information from an expanding receptive field.
In order to process auxiliary tokens, such as classification tokens, in convolutional layers, we reorganize input tokens $t$ into patch tokens in image form ($t_p$) and auxiliary tokens ($t_a$) using the function $\mathcal{K}$, as follows:
\begin{equation}
t_{a}, t_{p}=\mathcal{K}(t) \quad \text{where}  \quad \mathcal{K}:\R^{C\times N} \mapsto \R^{C\times A}, \R^{C\times H\times W}.
\end{equation}
where $C$ represents the embedding size, $N$ is the number of tokens, $A$ is the number of auxiliary tokens, and $H$ and $W$ are the height and width of the patch-token grid.

The Foveation module leverages multi-resolution information by applying successive depthwise separable convolutions ($\mathcal{DSC}$) to patch tokens $T_{p}$ and pointwise convolutions ($\mathcal{PC}$) to auxiliary tokens $T_{a}$, as the latter are two-dimensional features. This is mathematically represented as follows:
\begin{equation}
\begin{split}
t_p^l = \mathcal{DSC}^{l-1}(t_p^{l-1}) ~~~~~~~~ \text{and} ~~~~~~~~
t_a^l = \mathcal{PC}^{l-1}(t_a^{l-1}).
\end{split}
\end{equation}
Here, $l$ refers to the layer and $\mathcal{DSC}$ divides a kernel into two independent kernels that perform depthwise and pointwise convolutions, $\mathcal{DSC}(x) = \mathcal{PC}(\mathcal{DC}(x))$. As a result, each successive layer of the Foveation module employs progressively larger receptive fields. In all experiments, we use three foveation levels ($L_{\mathrm{fov}}=3$): the input features and two successive convolutional outputs. This fixed scale count matches the three-way channel split below, so each level contributes $C/3$ channels and the concatenated query, key, and value streams each recover width $C$.

To embed multi-scale information into the query, key, and value features, we split the auxiliary and patch-related features of each layer into three equal-sized splits in the channel dimension using function $\mathcal{S}$:
%
\begin{equation}
\begin{split}
t_{pq}, t_{pk}, t_{pv} &= \mathcal{S}(t_{p}), \\
\mathcal{S}: \R^{C \times H \times W} \;&\mapsto\;
\Big(\R^{C/3 \times H \times W}, \R^{C/3 \times H \times W}, \R^{C/3 \times H \times W}\Big), \\[2mm]
t_{aq}, t_{ak}, t_{av} &= \mathcal{S}(t_{a}), \\
\mathcal{S}: \R^{C \times A} \;&\mapsto\;
\Big(\R^{C/3 \times A}, \R^{C/3 \times A}, \R^{C/3 \times A}\Big).
\end{split}
\end{equation}

Finally, we merge and concatenate the auxiliary and patch splits from the three foveation levels in the channel dimension to form foveated-query ($q_{\mathrm{fov}}$), -key ($k_{\mathrm{fov}}$), and -value ($v_{\mathrm{fov}}$) features using the inverse of the operation that was applied for initial rearrangement:
%

\begin{equation}
\begin{split}
q_{\mathrm{fov}} &= [
\mathcal{K}^{-1}(t_{aq}^{0}, t_{pq}^{0}) \mid
\mathcal{K}^{-1}(t_{aq}^{1}, t_{pq}^{1}) \mid
\mathcal{K}^{-1}(t_{aq}^{2}, t_{pq}^{2})
],\\[1mm]
k_{\mathrm{fov}} &= [
\mathcal{K}^{-1}(t_{ak}^{0}, t_{pk}^{0}) \mid
\mathcal{K}^{-1}(t_{ak}^{1}, t_{pk}^{1}) \mid
\mathcal{K}^{-1}(t_{ak}^{2}, t_{pk}^{2})
],\\[1mm]
v_{\mathrm{fov}} &= [
\mathcal{K}^{-1}(t_{av}^{0}, t_{pv}^{0}) \mid
\mathcal{K}^{-1}(t_{av}^{1}, t_{pv}^{1}) \mid
\mathcal{K}^{-1}(t_{av}^{2}, t_{pv}^{2})
].
\end{split}
\end{equation}

\begin{figure}
    \centering
    \includegraphics[width=0.58\columnwidth]{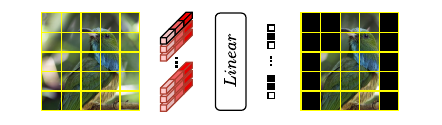}
    \caption{\textbf{Fixation module} utilizes multi-resolution information from the foveated-query token $q_{\mathrm{fov}}$ through a single linear layer, generating logits for the binary decision of identifying fixation points. The resulting fixation map then filters the foveated features for processing within the MHSA module.}
    \label{fig:fixation}
\end{figure}

\subsection{Fixation Module}

Humans execute a sequence of eye movements to construct a detailed scene map, selecting fixation points based on multi-resolution information from foveated perception. Inspired by this principle, we introduce a fixation module that leverages multi-resolution features to determine which input tokens should be processed.

Our fixation module uses the multi-resolution information in the foveated-query token $q_{\mathrm{fov}}$ to identify fixation points. This token's extensive receptive field allows fixation decisions to be based on a single token. We implement the module as a single linear layer that maps each foveated query to two logits for binary fixation decisions.
%

\begin{equation}
\begin{aligned}
\ell_{\mathrm{fix}}^{i,j} &= \operatorname{Linear}(q_{\mathrm{fov}}^{i,j}), \\
&\quad \forall i \in \{0, \dots, H-1\},\
       \forall j \in \{0, \dots, W-1\}.
\end{aligned}
\end{equation}

To generate a fixation map, we feed all the foveated query features to the fixation module (FIX). The fixation map is generated based on the logits produced by the module. The position where the first logit value is higher than the second logit value is set to 1.

\begin{equation}
\label{eq:gate}
g(i,j)=\begin{cases}
1, & \ell_{\mathrm{fix}}^{i,j}(0)>\ell_{\mathrm{fix}}^{i,j}(1),\\
0, & otherwise.
\end{cases}
\end{equation}

The fixation map is used to filter the foveated features that are processed in the Multi-Head Self-Attention (MHSA) module. See \Cref{fig:fixation}.

\subsection{Overall Architecture}
The HVS uses foveation and fixation mechanisms to efficiently process information. Inspired by this, we developed the Foveated Dynamic Transformer (FDT), a variant of the standard vision transformer that integrates computational foveation and fixation modules into its architecture. FDT retains the use of MHSA and MLP modules but processes only fixated tokens. Data flow in FDT, denoted by layer norm (LN), is structured as:

\begin{equation}
\begin{split}
    q_{\mathrm{fov}}, k_{\mathrm{fov}}, v_{\mathrm{fov}} = \operatorname{FOV}(\operatorname{LN}(x)), ~~~~~ g = \operatorname{FIX}(q_{\mathrm{fov}}), \\
%
q_{\mathrm{fix}}, k_{\mathrm{fix}}, v_{\mathrm{fix}}, x_{\mathrm{fix}} = \left\{ q_{\mathrm{fov}}^{i,j}, k_{\mathrm{fov}}^{i,j}, v_{\mathrm{fov}}^{i,j}, x^{i,j}\mid g(i,j) = 1 \right\}.
\end{split}
\end{equation}

In this architecture, the MHSA module forms a global relationship among fixated tokens and produces an attention matrix for each input token, focusing on specific fixated values. This module adapts to varying input sizes by processing solely the fixated tokens (where $d_{h}$ denotes the number of heads and $\sigma$ represents the softmax function):

\begin{equation}
\begin{aligned}
\mathrm{MHSA}(q_{\mathrm{fix}},k_{\mathrm{fix}},v_{\mathrm{fix}}) &=\sigma(q_{\mathrm{fix}}k_{\mathrm{fix}}^{\top} / \sqrt{d_{h}}) \cdot v_{\mathrm{fix}}
\end{aligned}
\end{equation}

%
FDT takes a full-size token map and produces foveated tokens, but since only fixated tokens are processed, their output shape ($x_{\mathrm{fix}}$) does not align with the expected input size for subsequent blocks. To resolve this, we blend processed and unprocessed tokens, ensuring proper input for the next stages.

\begin{equation}
out^{i,j}=\begin{cases}
x_{\mathrm{fix}}^{i,j}, & \text{if} \quad g(i,j) = 1,\\
x^{i,j}, & \text{otherwise}.
\end{cases}
\end{equation}

\subsection{Training Dynamic Network}

In the FDT architecture, the fixation module acts as a gating network that applies a selection operation to the foveated tokens of the input image.
Variations in input images result in differing numbers of selected tokens, which complicates training with mini-batches.
To facilitate mini-batch training, all foveated tokens are fed into the MHSA module without applying fixation sampling.
To enable end-to-end training with a differentiable fixation map, we apply Gumbel-Softmax with hard labeling to the output of the foveation module during training, replacing the logit comparison described in \cref{eq:gate}. Taking $g$ as the output of Gumbel-softmax;

\begin{equation}
\begin{aligned}
x_{\mathrm{masked}} & =\operatorname{MHSA}_{\mathrm{masked}}(q_{\mathrm{fov}},k_{\mathrm{fov}},v_{\mathrm{fov}},g)+x,\\
x_{\mathrm{masked}} & =\operatorname{MLP}(\operatorname{LN}(x_{\mathrm{masked}}))+x_{\mathrm{masked}}.
\end{aligned}
\end{equation}

The fixation map is then utilized for the calculation of masked attention (where $\mathcal{C}$ is a large constant):
%

\begin{equation}
\begin{aligned}
\mathrm{MHSA}_{\mathrm{masked}}(q_{\mathrm{fov}},&k_{\mathrm{fov}},v_{\mathrm{fov}},g) \\
&= \sigma \Big(
        q_{\mathrm{fov}} k_{\mathrm{fov}}^{\top} / \sqrt{d_h}
        \;+\; \mathbb{M}
      \Big) \cdot v_{\mathrm{fov}}, \\[1mm]
\mathbb{M}
    = -\mathcal{C} \;\times\;
       &\Big(
       1 - flatten(g) \otimes flatten(g)
       \Big).
\end{aligned}
\end{equation}


\paragraph{Budget Constraint.}
In order to achieve high performance, the fixation module tends to assign all tokens as fixation points when there are no budget constraints. However, the ideal behavior of the fixation module should be to focus on the minimum number of tokens necessary for accurate prediction. To achieve this, we introduce a fixation budget constraint that forces the network to allocate a certain percentage of all tokens. The fixation budget loss is defined as the blockwise $\ell_2$ norm between the desired percentile and the mean of the fixation map to measure the deviation from the desired budget ($g$ is the fixation map,
$\beta \in (0, 1]$ denotes the desired fixation budget, i.e., the target fraction of tokens to be selected as fixation points in each block, where
$L$ is the number of blocks):

\begin{equation}
\mathcal{L}_{\mathrm{Budget}}=\frac{1}{L}\sum_{\ell=0}^{L-1}\left\Vert \mathbb{E}(g)-\beta\right\Vert ^{2}.
\end{equation}


Therefore, the total loss for training FDT on a classification task is the sum of cross-entropy loss as the task loss and the fixation budget loss ($\lambda$ is the balancing factor):

\begin{equation}
\mathcal{L}=\mathcal{L}_{\mathrm{CE}} + \lambda \cdot \mathcal{L}_{\mathrm{Budget}}.
\end{equation}

%% file: sec/4.tex
\section{Empirical Analyses}
We demonstrate that FDT outperforms DeiT in robustness against adversarial attacks, natural corruption, and shortcut learning. We investigate the impact of budget size on both robustness and computational efficiency and provide attention and gating visualizations to support our findings. For experimental details, see \hyperref[sec:settings]{Appendix D}.

\subsection{Adversarial Robustness}

\begin{figure}[t]
    \centering
    \includegraphics[width=0.74\columnwidth]{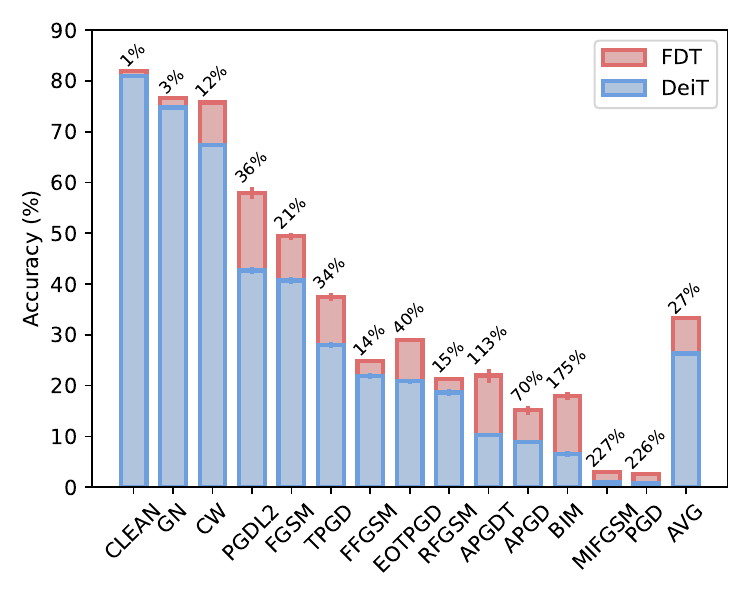}
    \caption{Adversarial robustness. FDT achieves a significant performance improvement, with an average increase of 27\% compared to DeiT across all attack types, excluding clean-data accuracy.}
    \label{fig:robustness}
\end{figure}

Although adversarial attacks target neural networks specifically, the human visual system is naturally robust to noise, clutter, and distribution shift. Mechanisms like foveation and fixation help humans emphasize meaningful features while discounting irrelevant perturbations. Motivated by this, we evaluated FDT using 12 adversarial attacks and one Gaussian-noise baseline. The results, shown in \cref{fig:robustness}, indicate that FDT outperformed the DeiT method in all types of attacks (for numerical results, see \cref{tab:adversarial_attack_all_models}).
We further provide a qualitative visualization of adversarial examples in \cref{fig:pgd_comparison}, where FDT perturbations appear more concentrated on semantically relevant image regions than DeiT perturbations.

%

\subsection{Natural Corruption Robustness}

\begin{table}[H]
\centering
\footnotesize
\setlength{\tabcolsep}{1.7pt}
\caption{Robustness to natural corruptions. Models are evaluated across five severity levels using the ImageNet100-C dataset. Values report mean top-1 accuracy (\%) for DeiT-S and FDT-S$_{0.5}$; full mean and standard-deviation results are provided in the Appendix.}
\label{tab:natural_corruption}
\begin{tabular}{ll|ccccc|ccccc}
\toprule
 & & \multicolumn{5}{c|}{DeiT-S} & \multicolumn{5}{c}{FDT-S$_{0.5}$} \\
\cmidrule(lr){3-7}\cmidrule(lr){8-12}
Category & Corruption & S1 & S2 & S3 & S4 & S5 & S1 & S2 & S3 & S4 & S5 \\
\midrule
\multirow{4}{*}{Blur}
 & Defocus Blur & 63.5 & 56.7 & 44.0 & 34.0 & 25.9 & 64.9 & 57.1 & 41.8 & 29.9 & 20.9 \\
 & Glass Blur & 68.8 & 60.7 & 46.0 & 39.5 & 32.2 & 70.6 & 63.2 & 48.0 & 41.0 & 32.4 \\
 & Motion Blur & 70.3 & 62.9 & 53.2 & 43.2 & 37.3 & 72.0 & 64.2 & 53.7 & 43.2 & 36.3 \\
 & Zoom Blur & 63.0 & 57.4 & 53.7 & 49.7 & 45.6 & 62.6 & 56.0 & 51.8 & 47.6 & 42.9 \\
\midrule
\multirow{4}{*}{Digital}
 & Contrast & 72.3 & 69.0 & 62.7 & 44.4 & 19.5 & 73.3 & 68.9 & 61.5 & 37.0 & 14.6 \\
 & Elastic Transform & 75.2 & 67.2 & 74.7 & 72.1 & 62.5 & 76.9 & 67.0 & 76.2 & 74.0 & 64.7 \\
 & JPEG Compression & 67.7 & 63.9 & 60.4 & 51.1 & 39.1 & 71.7 & 68.8 & 66.3 & 58.3 & 47.1 \\
 & Pixelate & 77.3 & 76.3 & 71.5 & 59.9 & 48.6 & 78.8 & 78.1 & 73.7 & 63.8 & 52.3 \\
\midrule
\multirow{3}{*}{Noise}
 & Gaussian Noise & 74.3 & 69.1 & 57.2 & 38.3 & 16.1 & 76.1 & 70.9 & 60.3 & 42.0 & 18.2 \\
 & Impulse Noise & 71.8 & 63.8 & 55.8 & 35.0 & 15.8 & 73.9 & 66.9 & 59.2 & 39.0 & 18.0 \\
 & Shot Noise & 74.2 & 67.6 & 55.4 & 32.7 & 18.4 & 76.1 & 69.7 & 58.8 & 36.4 & 21.1 \\
\midrule
\multirow{4}{*}{Weather}
 & Brightness & 78.8 & 77.2 & 75.2 & 71.3 & 65.5 & 80.1 & 78.7 & 77.3 & 74.1 & 69.3 \\
 & Fog & 68.0 & 62.4 & 53.5 & 49.4 & 38.3 & 70.2 & 64.8 & 56.0 & 52.4 & 43.1 \\
 & Frost & 73.2 & 66.5 & 59.8 & 59.2 & 54.1 & 75.8 & 69.3 & 63.2 & 62.2 & 57.0 \\
 & Snow & 66.9 & 52.0 & 53.7 & 44.7 & 41.9 & 69.0 & 55.4 & 57.9 & 48.8 & 46.6 \\
\midrule
\multicolumn{2}{l|}{Average} & 71.0 & 64.8 & 58.5 & 48.3 & 37.4 & 72.8 & 66.6 & 60.4 & 50.0 & 39.0 \\
\bottomrule
\end{tabular}
\end{table}

We evaluated models using the ImageNet100-C \citep{hendrycks2019benchmarkingneuralnetworkrobustness} dataset, which includes common corruptions. ImageNet100-C denotes ImageNet-C restricted to the same 100-class ImageNet100 subset used for clean training and validation. \Cref{tab:natural_corruption} shows that FDT consistently outperforms DeiT across all severity levels, particularly in weather-related corruptions. This robustness is consistent with the HVS-inspired motivation for adaptive, selective processing.

\subsection{Effect of Budget}

Our model incorporates the MHSA module, which exhibits quadratic growth in computational complexity as the number of tokens increases. The fixation module within each FDT block dynamically selects a subset of tokens for processing in the MHSA based on the input image's complexity, significantly reducing the overall computational cost.

To evaluate our model's efficiency, we quantified the Multiply-Accumulate operations (MACs) required for inference for both DeiT and FDT under various gating budgets. We calculated FDT's computational complexity by averaging the computations needed per sample in the validation set, reflecting the dynamic nature of our approach. MACs are normalized to DeiT to highlight computational efficiency gains.
\Cref{tab:complexity} shows that FDT requires fewer expected MACs for inference than DeiT. At a 50\% budget, FDT uses an average of 3.01 GMACs, achieving a 34.57\% reduction compared to DeiT. Even at a full budget, where all tokens are utilized, the increase in complexity is only 0.7\%, demonstrating the minimal overhead of the foveation and fixation modules.
These results indicate that FDT can dynamically allocate computational resources to more informative regions of the image, following an HVS-inspired selective-processing principle. At the 50\% budget, this allocation corresponds to higher clean accuracy and a 34.57\% MAC reduction relative to DeiT-S. The robustness and shortcut-learning results are consistent with the hypothesis that targeted processing can reduce reliance on irrelevant or misleading features.
These findings suggest that FDT can trade off clean accuracy, robustness metrics, and expected MACs across budgets, making it a candidate for future study in resource-constrained vision systems.
Furthermore, \Cref{tab:complexity} also serves as an ablation study: FDT$_{1.0}$(full budget) isolates the effect of the foveation module, while models with lower budgets highlight the contribution of the fixation mechanism to performance and robustness.

\begin{table}[!htbp]
\centering
\footnotesize
\setlength{\tabcolsep}{3.0pt}
\caption{Effect of gating budgets on computational efficiency and model performance. We report average accuracies for clean (Acc.), adversarially attacked (Adv.), and naturally corrupted (Corr.) samples, alongside clean accuracy for models trained on tinted samples (Shct.) as a robustness measure against shortcut learning. `Eff. Fix.' denotes the effective fixation ratio on the validation dataset, with subscripts indicating FDT's fixation budget. The Relative Gain Acc. column is the unweighted mean of per-axis percentage gains over DeiT, $\frac{1}{4}\sum_{a\in\{\mathrm{Acc.},\mathrm{Adv.},\mathrm{Corr.},\mathrm{Shct.}\}}100(a_{\mathrm{FDT}}-a_{\mathrm{DeiT}})/a_{\mathrm{DeiT}}$; Relative Gain GMAC is $100(\mathrm{GMAC}_{\mathrm{FDT}}-\mathrm{GMAC}_{\mathrm{DeiT}})/\mathrm{GMAC}_{\mathrm{DeiT}}$.}
\label{tab:complexity}
\begin{tabular}{|l|cccc|cc|cc|}
\cline{2-9}
\multicolumn{1}{c|}{} & \multicolumn{4}{c|}{Accuracy} & \multicolumn{2}{c|}{} & \multicolumn{2}{c|}{Relative Gain}\\
\multicolumn{1}{c|}{} & Acc. & Adv. & Corr. & Shct. & Eff. Fix. & GMAC & Acc. & GMAC\\
\hline
DeiT & 80.9{\tiny{}\textpm 0.11} & 26.3{\tiny{}\textpm 0.43} & 56.0{\tiny{}\textpm 0.18} & 53.5{\tiny{}\textpm 0.16} & 1.00{\tiny{}\textpm 0.000} & 4.60{\tiny{}\textpm 0.00} & 0 & 0\\ \hline
FDT\textsubscript{0.2} & 75.4{\tiny{}\textpm 2.55} & \textbf{45.9}{\tiny{}\textpm 0.40} & 48.2{\tiny{}\textpm 2.40} & 44.8{\tiny{}\textpm 1.07} & 0.29{\tiny{}\textpm 0.002} & 2.09{\tiny{}\textpm 0.01} & +9.38 & \textminus 54.57\\
FDT\textsubscript{0.3} & 80.0{\tiny{}\textpm 0.06} & 40.1{\tiny{}\textpm 1.71} & 54.5{\tiny{}\textpm 0.28} & 51.4{\tiny{}\textpm 0.56} & 0.37{\tiny{}\textpm 0.004} & 2.36{\tiny{}\textpm 0.01} & +11.19 & \textminus 48.70\\
FDT\textsubscript{0.4} & 81.5{\tiny{}\textpm 0.23} & 36.8{\tiny{}\textpm 3.42} & 56.6{\tiny{}\textpm 0.26} & 54.0{\tiny{}\textpm 0.29} & 0.45{\tiny{}\textpm 0.007} & 2.63{\tiny{}\textpm 0.03} & +10.67 & \textminus 42.83\\
FDT\textsubscript{0.5} & 81.9{\tiny{}\textpm 0.05} & 33.3{\tiny{}\textpm 0.45} & 57.8{\tiny{}\textpm 0.17} & 56.5{\tiny{}\textpm 0.53} & 0.56{\tiny{}\textpm 0.006} & 3.01{\tiny{}\textpm 0.02} & +9.17 & \textminus 34.57\\
FDT\textsubscript{0.6} & 82.8{\tiny{}\textpm 0.19} & 32.8{\tiny{}\textpm 0.52} & 59.2{\tiny{}\textpm 0.38} & 57.9{\tiny{}\textpm 0.29} & 0.68{\tiny{}\textpm 0.003} & 3.47{\tiny{}\textpm 0.01} & +10.25 & \textminus 24.57\\
FDT\textsubscript{0.7} & 83.1{\tiny{}\textpm 0.17} & 31.8{\tiny{}\textpm 0.23} & 59.9{\tiny{}\textpm 0.30} & 59.5{\tiny{}\textpm 0.19} & 0.81{\tiny{}\textpm 0.004} & 3.93{\tiny{}\textpm 0.01} & +10.45 & \textminus 14.57\\
FDT\textsubscript{0.8} & 84.0{\tiny{}\textpm 0.42} & 31.5{\tiny{}\textpm 0.09} & 60.9{\tiny{}\textpm 0.40} & 61.1{\tiny{}\textpm 0.34} & 0.93{\tiny{}\textpm 0.001} & 4.36{\tiny{}\textpm 0.01} & +11.64 & \textminus 5.22\\
FDT\textsubscript{0.9} & 83.9{\tiny{}\textpm 0.18} & 32.4{\tiny{}\textpm 0.08} & 61.8{\tiny{}\textpm 0.40} & 61.8{\tiny{}\textpm 0.72} & 0.99{\tiny{}\textpm 0.001} & 4.59{\tiny{}\textpm 0.00} & +13.19 & \textminus 0.22\\
FDT\textsubscript{1.0} & \textbf{84.5}{\tiny{}\textpm 0.07} & 32.5{\tiny{}\textpm 0.68} & \textbf{62.9}{\tiny{}\textpm 0.17} & \textbf{62.4}{\tiny{}\textpm 0.18} & 1.00{\tiny{}\textpm 0.000} & 4.63{\tiny{}\textpm 0.00} & +14.25 & +0.65\\
\hline
\end{tabular}
\end{table}

\subsection{Robustness against Shortcut Learning}
%
%
\begin{figure}[t]
    \centering
    \includegraphics[width=0.82\columnwidth]{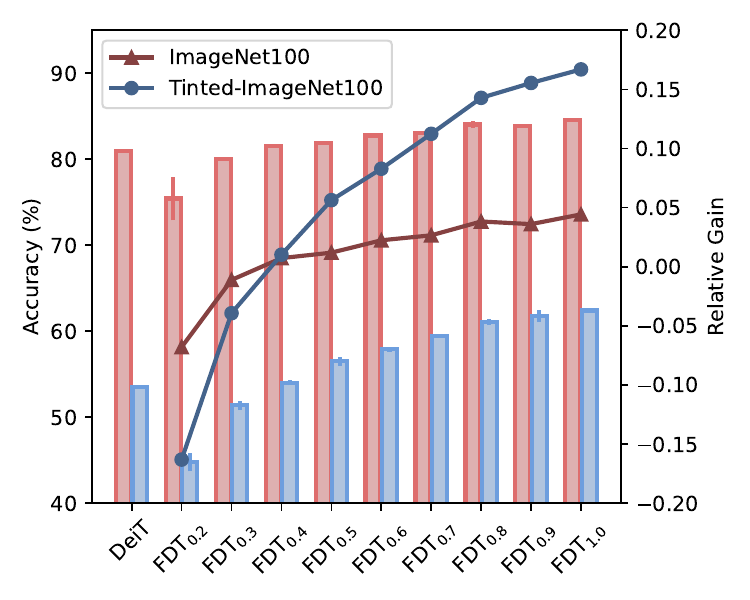}
    \caption{Robustness to shortcut learning. Actual performance of each model is displayed as bars, with the relative gains of FDT over DeiT depicted as a line graph.}
    \label{fig:tinted_experiments}
\end{figure}

Shortcut learning occurs when ANNs form decision rules that excel on specific datasets by exploiting spurious correlations or statistical irregularities instead of learning the underlying task. These strategies, though effective on familiar data, do not generalize well across different data distributions \citep{geirhos2020shortcut}.
To assess models, we trained both DeiT and the FDT under varying budget constraints using the Tinted-ImageNet100 dataset, where each training sample is modified with a class-specific tint, following the approach in the Tinted-STL10 dataset. Shortcut-learning performance is evaluated on the untinted standard ImageNet100 validation set, so the reported accuracy measures whether models generalize without the tint cue. \Cref{fig:tinted_experiments} suggests that FDT is more robust to this synthetic shortcut than DeiT, especially under larger budgets.

\subsection{Feature Inversion}

We employ ``feature inversion" to enhance our understanding and visualization of transformer-based representations. This technique reconstructs an input image from specific model features or activations, offering insights into how the model processes inputs and makes predictions. While widely used with CNNs \citep{simonyan2013deep, selvaraju2017grad}, its application to transformers remains less explored.
Using the `Deep Image Prior' image parameterization of \citet{ulyanov2018deep} and the feature-inversion setup of \citet{engstrom2019adversarial}, we optimize a CNN-based generator $F_\theta(\cdot)$ to transform random noise input $z$ into an image. The goal is to match the features of the output image, particularly the $\mathrm{CLS}$ features, with those of a target image $I$, defined by:

\begin{equation}
    \underset{\theta}{\arg \min }\left\|\phi\left(F_\theta(z)\right)-\phi(I)\right\|_{\mathrm{F}}
    \label{eq:inverted_features}
\end{equation}

where $\phi(I)$ denotes the target features and $\|\cdot\|_{\mathrm{F}}$ represents the Frobenius norm, focusing on the classification token ($\mathrm{CLS}$); thus, $\phi(I)=\mathrm{CLS}(I)$.

\begin{figure}[t]
    \centering
    \includegraphics[width=\columnwidth]{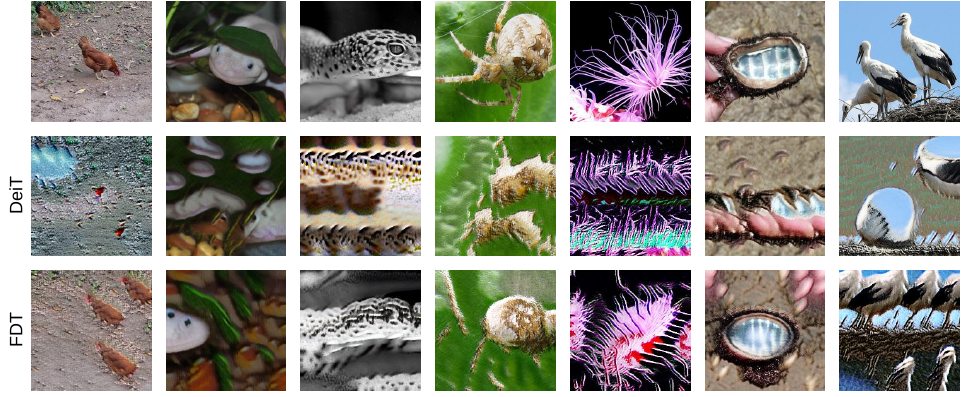}
    \caption{Feature inversion using the `Deep Image Prior' method for DeiT-S and FDT-S trained at a 50\% budget. This qualitative comparison suggests that FDT reconstructions preserve clearer object structure than DeiT reconstructions.}
    \label{fig:inverted_features}
\end{figure}

We employ the same network architecture and parameters as \citet{engstrom2019adversarial} for the generative network $F_\theta(z)$. The qualitative examples in \cref{fig:inverted_features} show clearer object structure in FDT reconstructions than in DeiT reconstructions, consistent with the hypothesis \citep{engstrom2019adversarial} that robustness-oriented models can yield more interpretable representations.


\subsection{Effect of Model Size}

The relationship between model size and performance is pivotal in neural network design, balancing capacity with computational efficiency. Generally, larger models can achieve higher accuracy but may overfit and require more resources. Smaller models might better generalize and suit practical needs but have limited learning capacity for certain functions. Understanding this trade-off is crucial for choosing the appropriate model size for specific tasks.

We explore the impact of model size on our FDT by evaluating three sizes: tiny, small, and base, following DeiT conventions.
Importantly, to ensure a fair and controlled comparison, we maintained the architectural backbone identical to DeiT, with the only differences being the addition of the foveation and fixation modules.
These additions are lightweight: for example, in the small model, FDT has 21.85 million parameters compared to DeiT-S's 21.70 million, with the foveation block replacing the 5.32 million parameter QKV linear layer with 5.47 million parameters (a marginal 0.15 million increase) and the fixation block adding just 9k parameters. This minimal parameter increase ensures that any performance gains can be attributed to the introduced modules rather than increased capacity.

We compared FDT's performance against DeiT across the same evaluation metrics, except for the learning rate of the base model set to 2e-4. Our results, presented in \cref{tab:model_size}, cover accuracy on clean, adversarially attacked, and naturally corrupted images, and the relative accuracy gains over DeiT.
These results suggest that FDT remains competitive across model sizes in the evaluated clean, adversarial, and corruption settings.

\begin{table}[H]
\centering
\footnotesize
\setlength{\tabcolsep}{4.0pt}
\caption{Comparison of FDT across different model sizes (tiny, small, and base) against the DeiT model in terms of accuracy on clean, adversarially attacked, and naturally corrupted images. Rel. Gain is the unweighted mean of per-axis percentage gains over the DeiT model of the same size, $\frac{1}{3}\sum_{a\in\{\mathrm{Acc.},\mathrm{Adv.},\mathrm{Corr.}\}}100(a_{\mathrm{FDT}}-a_{\mathrm{DeiT}})/a_{\mathrm{DeiT}}$.}
\label{tab:model_size}
\begin{tabular}{|l|l|c|ccc|c|}
\hline
Size                   & Model & GMAC                       & Acc.                      & Adv. Acc.                 & Corr. Acc.                & Rel. Gain \\ \hline
\multirow{2}{*}{Tiny}  & DeiT  & 1.26{\tiny{}\textpm 0.00}  & 64.4{\tiny{}\textpm 0.48} & 17.6{\tiny{}\textpm 0.32} & 40.8{\tiny{}\textpm 0.24} & 0         \\
                       & FDT   & 0.84{\tiny{}\textpm 0.00}  & 67.1{\tiny{}\textpm 0.08} & 24.7{\tiny{}\textpm 1.67} & 43.5{\tiny{}\textpm 0.15} & +17.05    \\ \hline
\multirow{2}{*}{Small} & DeiT  & 4.60{\tiny{}\textpm 0.00}  & 80.9{\tiny{}\textpm 0.11} & 26.3{\tiny{}\textpm 0.43} & 56.0{\tiny{}\textpm 0.18} & 0         \\
                       & FDT   & 3.01{\tiny{}\textpm 0.02}  & 81.9{\tiny{}\textpm 0.05} & 33.3{\tiny{}\textpm 0.45} & 57.8{\tiny{}\textpm 0.17} & +10.36    \\ \hline
\multirow{2}{*}{Base}  & DeiT  & 17.57{\tiny{}\textpm 0.00} & 81.5{\tiny{}\textpm 0.25} & 30.0{\tiny{}\textpm 0.76} & 56.9{\tiny{}\textpm 0.47} & 0         \\
                       & FDT   & 11.56{\tiny{}\textpm 0.07} & 82.2{\tiny{}\textpm 0.28} & 40.8{\tiny{}\textpm 1.80} & 57.2{\tiny{}\textpm 0.31} & +12.46    \\ \hline
\end{tabular}
\end{table}

\begin{figure}[H]
    \centering
    \includegraphics[width=\columnwidth]{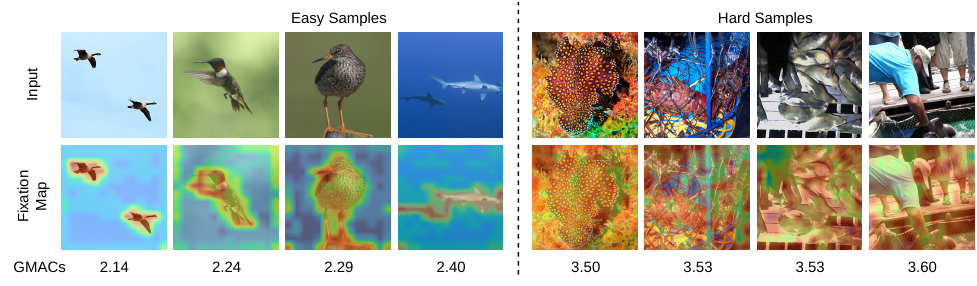}
    \caption{Visual comparison of easy and hard samples with their computational requirements for inference. The first row shows the original images, the second row shows heatmaps of fixation points, and the final row lists the GMACs required for each image. This highlights how computational load varies with the number and distribution of fixation points, reflecting inference efficiency and focus}
    \label{fig:easy_hard_samples}
\end{figure}

\subsection{Input-Dependent Compute Analysis}
Reaction-time (RT) in HVS measures the duration to respond to a visual stimulus, reflecting neural and cognitive processes in perception, attention, and decision-making, and varies with task complexity, object number and similarity, and uncertainties like occlusions or viewpoint changes, generally increasing with higher difficulty. We use the reaction-time analogy as a qualitative way to describe input-dependent computational load in FDT, not as a measured comparison to human response times. Because FDT dynamically selects fixation points, different images require different numbers of processed tokens and therefore different GMACs. To illustrate this behavior, we divide validation samples into balanced easy, medium, and hard groups according to their average number of fixations. \Cref{tab:hardness_acc} reports accuracy across these groups, and \cref{fig:easy_hard_samples} visualizes representative samples and their compute requirements. Ultimately, these results demonstrate that FDT naturally scales its computational effort in response to image complexity, mirroring the reaction-time dynamics of the human visual system.

\subsection{Model Decisions: Visualization and Insights}

\begin{figure}[H]
    \centering
    \includegraphics[width=0.88\columnwidth]{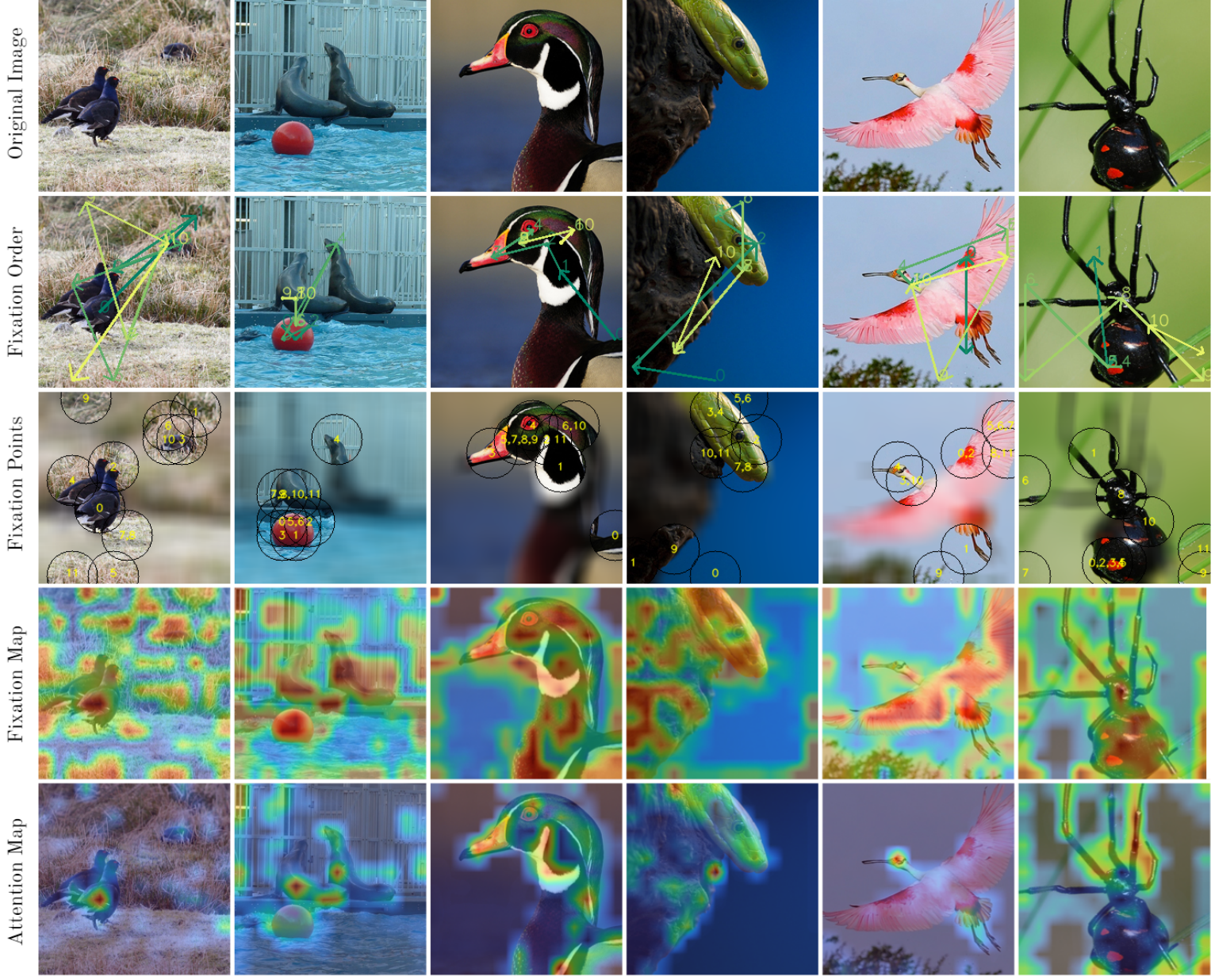}
    \caption{Visualization of fixation points, blockwise fixation order, fixation maps, and attention maps. Each column displays diverse samples from the ImageNet100 validation set. The order markers denote transformer block depth; circle sizes are fixed for legibility and do not encode fixation strength, probability, or uncertainty.}
    \label{fig:vis-fix-att}
\end{figure}

We visualize model fixation maps and attention maps as HVS-inspired diagnostics of selective processing. The FDT fixation module generates maps that highlight fixation probabilities across image tokens.
To illustrate the model's blockwise fixation selections, we identify the most likely token positions as fixation points. \Cref{fig:vis-fix-att} shows these points across transformer block depth: arrows and green-to-yellow color transitions indicate block order, and circles mark fixation points with a fixed display size chosen for legibility rather than to encode fixation strength or probability. Focus intensity adapts to scenes with multiple objects through diverse sampling. Tokens are selected via Gumbel-Softmax and hard label techniques, producing binary maps. Averaging maps across blocks yields overall fixation maps (\Cref{fig:vis-fix-att}), showing how the model prioritizes informative content while minimizing irrelevant background. Additionally, attention rollout \citep{abnar2020quantifying} visualizes focus on the classification token, revealing that while fixation maps cover broad informative areas, attention maps concentrate on discriminative regions for class identification. These visualizations qualitatively support the intended selective-processing behavior.

%% file: sec/6.tex
\section{Conclusion}

We introduced the \textit{Foveated Dynamic Transformer (FDT)}, a novel architecture inspired by human visual system mechanisms, enhancing computational efficiency and robustness against adversarial attacks, natural corruption, and shortcut learning. Our results on the ImageNet100 dataset show that, at selected budgets, FDT can achieve higher clean accuracy, lower expected MACs, and stronger robustness metrics than the baseline architecture. This research contributes to biologically inspired computational models, integrating human visual principles into deep learning architectures. The FDT balances high performance with computational efficiency in the evaluated settings, highlighting its HVS-inspired capability to focus computation on informative image regions. Future work could evaluate FDT in domains such as video processing and augmented reality, where dynamic foveation may reduce computational demands while maintaining performance, and explore its deployment in real-world, resource-limited scenarios, underscoring the value of bioinspired approaches in developing efficient and robust AI systems.

%% file: sec/5.tex
\section{Limitations}

Although FDT achieves substantial MAC savings relative to the baseline, reduced MACs do not necessarily translate to lower wall-clock latency: the foveation pipeline requires token re-arrangement, channel splits and concat/merge operations to form multiscale Q, K, V tensors, which introduce memory-bound overheads that can dominate runtime.
Our primary goal, however, is not to optimize latency, but to demonstrate how HVS-inspired design choices can improve robustness and compute efficiency. Inference latency ultimately depends more on platform and software stack than on MACs alone, and addressing these overheads through kernel-level optimizations or specialized implementations is left as future work to better align theoretical efficiency with measured latency.

FDT should also be interpreted as HVS-inspired rather than as a faithful biological model. In the current architecture, all spatial locations receive the same multi-scale feature construction, so there is no explicit peripheral degradation tied to retinal eccentricity. Multiple fixation regions are selected simultaneously within a single feedforward pass rather than through a temporal sequence of saccades, the model has no recurrence or mechanism for revisiting earlier locations, and the selected fixation set can change from block to block. These differences mean that foveation and fixation serve as design motivations for adaptive computation and token selection, not as claims of biological fidelity.

%% file: sec/X_supp.tex
\clearpage
\maketitlesupplementary

\setcounter{section}{0}


\section{Broader Impact}

The introduction of the Foveated Dynamic Transformer (FDT) represents a step toward more adaptive artificial vision systems, offering enhanced robustness and efficiency in the evaluated ImageNet100 settings. The FDT's robustness to adversarial attacks and natural corruption without explicit adversarial training could motivate future studies in safety-sensitive domains such as autonomous driving, medical imaging, and surveillance, where accuracy and reliability are critical. Additionally, the model's reduced expected MACs may support future deployment work in environments with limited computational resources. As an HVS-inspired design, FDT may foster interdisciplinary collaboration between AI researchers and neuroscientists while motivating future work on adaptive visual processing. Ultimately, the FDT's approach could inspire new research directions, advancing both artificial intelligence and cognitive science.

\section{Extended Related Work}
\label{sec:extended_related}

Researchers have also examined foveation from complementary computational and perceptual perspectives. \citet{deza2020emergent} investigated how foveated image transformations influence downstream visual representations in two-stage machine vision models, focusing on the role of texture-based peripheral coding. In a similar direction, \citet{thavamani2021fovea} introduced a foveated object detector that applies image magnification to preserve high-resolution detail at points of interest while maintaining a compact canvas size.

A separate line of work has advanced efficiency in transformer-based vision models through token slimming, pruning, and related reduction strategies. Several methods explore how to discard uninformative visual tokens while retaining task-relevant structure (\citet{mao2025efficienttokencompressionvision}; \citet{yu2023xprunerexplainablepruningvision}; \citet{rao2021dynamicvitefficientvisiontransformers}; \citet{kim2022learnedtokenpruningtransformers}; \citet{wang2021adaptivefocusefficientvideo}). Building on this trend, \citet{tang2022patchslimmingefficientvision} propose progressively removing less informative tokens during the forward pass to reduce computational overhead. \citet{wang2024zerotprunezeroshottokenpruning} introduce Zero-TPrune, a zero-shot token pruning framework that leverages structural properties of the transformer attention graph to identify redundant tokens without additional training. \citet{zhang2024synergistic} present a synergistic patch pruning approach that integrates both intra- and inter-layer importance signals to eliminate unnecessary patches across layers. Complementing these methods, \citet{Zheng_2023_ICCV} propose Focus-DETR, a DETR variant that employs dual attention and token scoring to emphasize informative regions and reduce inference cost while preserving detection accuracy.

Although these token-slimming approaches improve computational efficiency by selectively removing tokens or patches, they are driven primarily by engineering considerations and do not attempt to model visual attention from a biologically grounded perspective.

\section{Adversarial Attack Qualitative Analysis}

\begin{figure}[htbp]
\centering

\begingroup
\newcommand{\pgdlabel}[1]{\raisebox{0.82cm}[0pt][0pt]{\makebox[1.5cm][r]{#1}}}
\newcommand{\pgdimg}[1]{\includegraphics[width=2cm]{#1}}
\setlength{\tabcolsep}{0pt}
\begin{tabular}{@{}r@{\hspace{0.08cm}}l@{}}
\makebox[1.5cm][r]{} &
\pgdimg{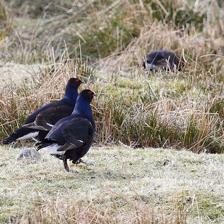}%
\pgdimg{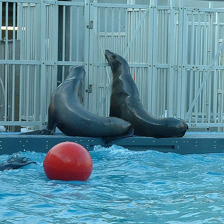}%
\pgdimg{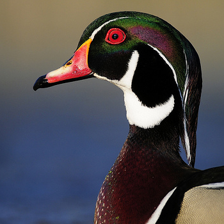}%
\pgdimg{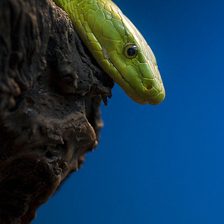}%
\pgdimg{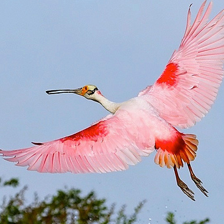}%
\pgdimg{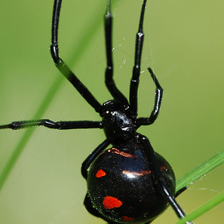}\tabularnewline
\pgdlabel{$FDT_{0.2}$} &
\pgdimg{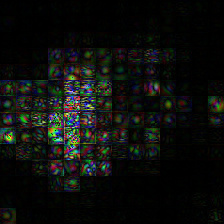}%
\pgdimg{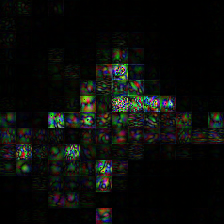}%
\pgdimg{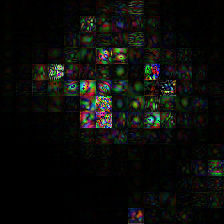}%
\pgdimg{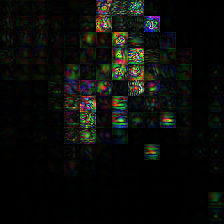}%
\pgdimg{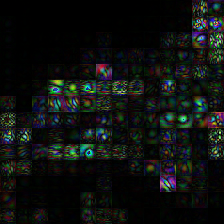}%
\pgdimg{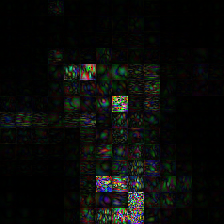}\tabularnewline
\pgdlabel{DeiT} &
\pgdimg{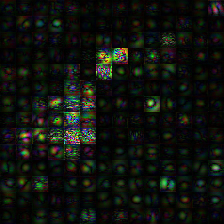}%
\pgdimg{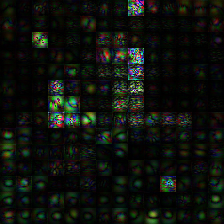}%
\pgdimg{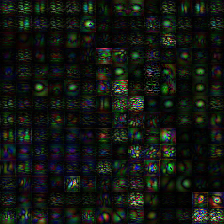}%
\pgdimg{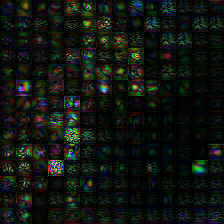}%
\pgdimg{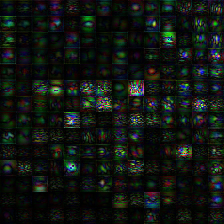}%
\pgdimg{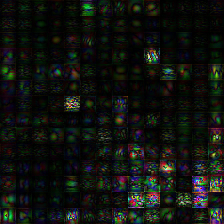}\tabularnewline
\pgdlabel{$FDT_{0.2}$} &
\pgdimg{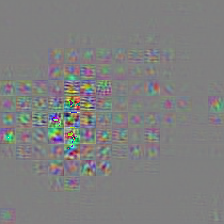}%
\pgdimg{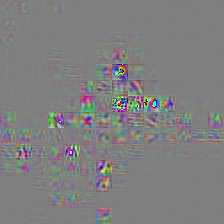}%
\pgdimg{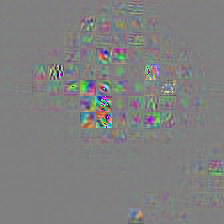}%
\pgdimg{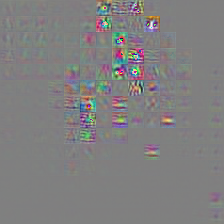}%
\pgdimg{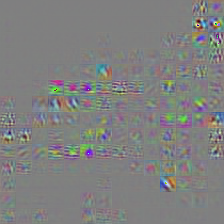}%
\pgdimg{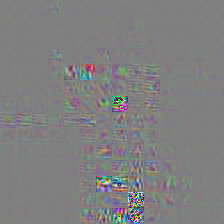}\tabularnewline
\pgdlabel{DeiT} &
\pgdimg{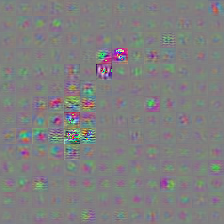}%
\pgdimg{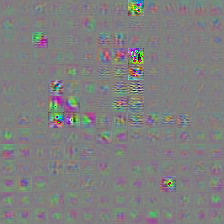}%
\pgdimg{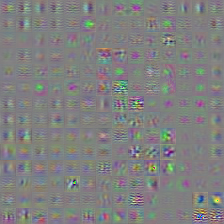}%
\pgdimg{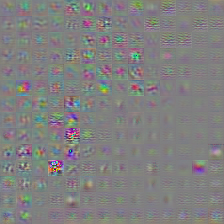}%
\pgdimg{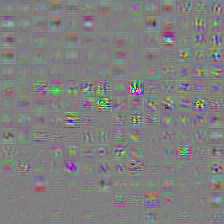}%
\pgdimg{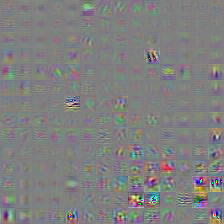}
\end{tabular}
\endgroup

\caption{Visualization of perturbations generated by the PGD-L2 attack. The second and third rows show the absolute values of the attack pattern. The fourth and fifth rows show the shifted perturbation images, where all values are shifted by 128 to properly present both positive and negative values. For better visibility, values are multiplied by 20 for the shifted images and by 50 for the absolute value images.}
\label{fig:pgd_comparison}
\end{figure}

To complement the quantitative evaluation, a visual comparison is provided between adversarial examples generated for the FDT model and those for the DeiT baseline. This qualitative analysis offers a more intuitive understanding of the differences in robustness between the two models.

Figure~\ref{fig:pgd_comparison} illustrates the perturbation patterns produced using the PGD-L2 attack with $\epsilon=1$ over 100 optimization steps. This PGD-L2 visualization is separate from the $\ell_\infty$ PGD evaluation row, which uses $\epsilon=0.1$ as reported in \cref{tab:attack_hparams}. The figure includes multiple visualization strategies to enhance interpretability. The second and third rows display the absolute values of the perturbations, scaled by a factor of 50 to improve visibility. The fourth and fifth rows present shifted versions of the perturbation maps, where pixel values are offset by 128, a standard approach for visualizing both positive and negative perturbation magnitudes. These shifted images are scaled by a factor of 20.

In this qualitative visualization, perturbations for FDT appear more concentrated in semantically relevant image regions, whereas those for DeiT appear more diffuse and less structured. This pattern is consistent with the quantitative robustness results, but it should be read as illustrative rather than as a separate robustness metric.



\begin{table}[htbp]
\centering
\begingroup
\footnotesize
\setlength{\tabcolsep}{3.0pt}
\renewcommand{\arraystretch}{0.95}
\begin{tabular}{|l|l|l|ccc|}
\cline{4-6}
\multicolumn{1}{l}{} & \multicolumn{1}{l}{} &  & Easy  & Medium  & Hard \\
\hline
\multicolumn{3}{|l|}{Clean} & 84.9{\tiny{}\textpm 0.35} & 80.5{\tiny{}\textpm 0.25} & 80.3{\tiny{}\textpm 0.31} \\
\hline
\multicolumn{2}{|l|}{Adversarial Att.} & PGD $\ell_2$  & 60.6{\tiny{}\textpm 0.35} & 56.4{\tiny{}\textpm 2.12} & 56.6{\tiny{}\textpm 1.32} \\
\hline
\multirow{15}{*}{\rotatebox[origin=c]{90}{Natural Corruption}}
 & \multirow{4}{*}{Blur} & Defocus B.  & 45.9{\tiny{}\textpm 0.81} & 43.8{\tiny{}\textpm 0.72} & 39.1{\tiny{}\textpm 0.94} \\
 &  & Glass B.  & 55.7{\tiny{}\textpm 0.71} & 52.0{\tiny{}\textpm 0.06} & 45.5{\tiny{}\textpm 0.07} \\
 &  & Motion B.  & 60.0{\tiny{}\textpm 0.24} & 54.4{\tiny{}\textpm 0.36} & 47.2{\tiny{}\textpm 1.33} \\
 &  & Zoom B.  & 63.0{\tiny{}\textpm 1.33} & 52.1{\tiny{}\textpm 0.24} & 41.4{\tiny{}\textpm 0.64} \\
\cline{2-6}
 & \multirow{4}{*}{Digital} & Contrast  & 45.0{\tiny{}\textpm 1.26} & 51.1{\tiny{}\textpm 1.20} & 57.1{\tiny{}\textpm 0.53} \\
 &  & Elastic Tran.  & 75.4{\tiny{}\textpm 0.31} & 71.4{\tiny{}\textpm 0.37} & 68.5{\tiny{}\textpm 0.44} \\
 &  & JPEG Comp.  & 68.7{\tiny{}\textpm 1.31} & 62.2{\tiny{}\textpm 0.77} & 56.4{\tiny{}\textpm 0.65} \\
 &  & Pixelate  & 74.8{\tiny{}\textpm 0.28} & 68.4{\tiny{}\textpm 0.51} & 64.8{\tiny{}\textpm 0.72} \\
\cline{2-6}
 & \multirow{3}{*}{Noise} & Gaussian N.  & 54.4{\tiny{}\textpm 1.20} & 52.2{\tiny{}\textpm 1.07} & 54.0{\tiny{}\textpm 1.59} \\
 &  & Impulse N.  & 53.0{\tiny{}\textpm 1.65} & 50.1{\tiny{}\textpm 1.40} & 51.2{\tiny{}\textpm 1.37} \\
 &  & Shot N.  & 53.8{\tiny{}\textpm 1.44} & 50.9{\tiny{}\textpm 1.37} & 52.5{\tiny{}\textpm 1.69} \\
\cline{2-6}
 & \multirow{4}{*}{Weather} & Brightness  & 79.3{\tiny{}\textpm 0.77} & 75.0{\tiny{}\textpm 0.10} & 73.4{\tiny{}\textpm 0.38} \\
 &  & Fog  & 61.1{\tiny{}\textpm 0.85} & 57.4{\tiny{}\textpm 1.00} & 53.5{\tiny{}\textpm 0.35} \\
 &  & Frost  & 70.6{\tiny{}\textpm 0.54} & 64.3{\tiny{}\textpm 0.95} & 61.7{\tiny{}\textpm 0.43} \\
 &  & Snow  & 63.0{\tiny{}\textpm 0.46} & 54.6{\tiny{}\textpm 0.40} & 49.1{\tiny{}\textpm 1.20} \\
\hline
\end{tabular}
\endgroup
\caption{Accuracy metrics across samples of varying difficulty levels, subjected to different noise types. The validation dataset was divided into three balanced subsets based on average fixations per sample: easy' for the fewest, medium' for an intermediate number, and `hard' for the most fixations. For analyses of natural corruption, samples from all severity levels were included.}
\label{tab:hardness_acc} %
\end{table}

\section{Experimental Settings}\label{sec:settings}

To evaluate FDT, we conduct image classification experiments on the ImageNet-100 dataset, a subset of the ImageNet-1k dataset containing 100 randomly chosen classes with 1300 training images and 50 validation images per class. We report top-1 accuracy using a single 224x224 crop and compare FDT to DeiT \citep{touvron2021training} using the same training settings, including AdamW optimization for 300 epochs with cosine learning rate decay and 20 epochs of linear warm-up, batch size of 512, and V100 GPUs for training and evaluation. The initial learning rate, weight decay, and momentum are set to 0.001, 0.05, and 0.9, respectively. Unless otherwise stated, all experiments use a budget of $\beta=0.5$. All results are reported as the mean and one standard deviation of three differently initialized runs.
For training Tiny, Small, and Base models, one, two, and four V100 GPUs with 32GB of memory are used, respectively. Evaluations of the models are done on a single V100 GPU.

\paragraph{\textbf{Adversarial Robustness.}}
The robustness evaluation includes a Gaussian-noise (GN) baseline and 12 adversarial attacks: CW \citep{carlini2017towards}, PGD-L2 and PGD-$\ell_\infty$ \citep{madry2017towards}, FGSM \citep{goodfellow2014explaining}, TPGD \citep{zhang2019theoretically}, FFGSM \citep{wong2020fast}, EOTPGD \citep{liu2018adv}, RFGSM \citep{tramer2017ensemble}, APGD and APGDT \citep{croce2020reliable}, BIM \citep{kurakin2018adversarial}, and MIFGSM \citep{dong2018boosting}. We used TorchAttacks v3.3.0 \citep{kim2020torchattacks}, the latest public release available on Nov. 17, 2022. Unless otherwise noted in \cref{tab:attack_hparams}, attack arguments follow the TorchAttacks v3.3.0 defaults. The row labeled PGD corresponds to the CSV column PGD0.1 and uses an $\ell_\infty$ budget of $\epsilon=0.1$; PGDL2 is a distinct $\ell_2$ attack with $\epsilon=1.0$. All adversarial examples are generated against the inference-time FDT forward pass: the fixation map is computed by the deterministic FIX logit comparison in \cref{eq:gate}, selected tokens are processed and merged back into the token stream, and the Gumbel-Softmax masked-attention surrogate used during training is not used for attack evaluation. As demonstrated in Table \ref{tab:adversarial_attack_all_models} in the Appendix, although a lower budget results in some accuracy drops, it overall helps to produce a more robust model. These findings support that the FDT method effectively enhances robustness against adversarial attacks by focusing on relevant features and filtering out noise.

\newpage

\refstepcounter{section}\phantomsection
\addcontentsline{toc}{section}{\Alph{section}. Adversarial Attack Results}
\noindent{\Large\bfseries \Alph{section}. Adversarial Attack Results}\par\vspace{1em}

\par\noindent\refstepcounter{table}\textbf{Table \thetable}\label{tab:attack_hparams}\par
\noindent Adversarial-evaluation hyperparameters. All attacks were implemented with TorchAttacks v3.3.0. `--' indicates that the field is not applicable.\par\vspace{0.5em}
\begin{center}
\begingroup
\small
\setlength{\tabcolsep}{3.0pt}
\resizebox{\textwidth}{!}{%
\begin{tabular}{llllll}
\toprule
Evaluation row & Norm & Budget & Steps & Step size & Restarts / notes \\
\midrule
GN & -- & $\sigma=0.1$ & -- & -- & Gaussian noise; clipped to $[0,1]$ \\
FGSM & $\ell_\infty$ & $\epsilon=8/255$ & 1 & -- & single gradient-sign step \\
BIM & $\ell_\infty$ & $\epsilon=8/255$ & 10 & $\alpha=2/255$ & no random restart \\
CW & $\ell_2$ objective & -- & 50 & lr $=0.01$ & $c=1$, $\kappa=0$; no binary search \\
RFGSM & $\ell_\infty$ & $\epsilon=8/255$ & 10 & $\alpha=2/255$ & random sign initialization \\
PGD & $\ell_\infty$ & $\epsilon=0.1$ & 5 & $\alpha=2/255$ & random start; overrides default $\epsilon=8/255$ and 10 steps \\
PGDL2 & $\ell_2$ & $\epsilon=1.0$ & 5 & $\alpha=0.2$ & random start; Fig.~\ref{fig:pgd_comparison} uses 100 steps \\
EOTPGD & $\ell_\infty$ & $\epsilon=8/255$ & 1 & $\alpha=2/255$ & random start; EOT iterations $=2$ \\
TPGD & $\ell_\infty$ & $\epsilon=8/255$ & 3 & $\alpha=2/255$ & KL-divergence objective \\
FFGSM & $\ell_\infty$ & $\epsilon=8/255$ & 1 & $\alpha=10/255$ & random uniform initialization \\
MIFGSM & $\ell_\infty$ & $\epsilon=8/255$ & 10 & $\alpha=2/255$ & momentum decay $=1.0$ \\
APGD & $\ell_\infty$ & $\epsilon=8/255$ & 10 & adaptive & one restart; seed $=0$, CE loss, EOT $=1$, $\rho=0.75$ \\
APGDT & $\ell_\infty$ & $\epsilon=8/255$ & 10 & adaptive & one restart; seed $=0$, targeted DLR, EOT $=1$, $\rho=0.75$ \\
\bottomrule
\end{tabular}}
\endgroup
\end{center}

\par\noindent\refstepcounter{table}\textbf{Table \thetable}\label{tab:adversarial_attack_all_models}\par
\noindent Comparison of robustness against 12 adversarial attacks and one Gaussian-noise baseline for FDT and DeiT models trained on ImageNet-100 dataset. The models are labeled with their corresponding fixation budget hyperparameter (subscript) and model size (T, S, B for tiny, small, and base, respectively). The mean and one standard deviation of three runs with different initializations are reported.\par\vspace{0.5em}
\begin{center}
\begin{small}
\setlength{\tabcolsep}{1.7pt}
\resizebox{\textwidth}{!}{%
\begin{tabular}{|l|ccc|ccccccccccc|}
\hline
Attack Type & DeiT-T & DeiT-S & DeiT-B & FDT-T$_{0.5}$ & FDT-S$_{0.2}$ & FDT-S$_{0.3}$ & FDT-S$_{0.4}$ & FDT-S$_{0.5}$ & FDT-S$_{0.6}$ & FDT-S$_{0.7}$ & FDT-S$_{0.8}$ & FDT-S$_{0.9}$ & FDT-S$_{1.0}$ & FDT-B$_{0.5}$\tabularnewline
\hline
CLEAN & 64.4{\tiny{}\textpm 0.48} & 80.9{\tiny{}\textpm 0.11} & 81.5{\tiny{}\textpm 0.25} & 67.1{\tiny{}\textpm 0.08} & 75.4{\tiny{}\textpm 2.55} & 80.0{\tiny{}\textpm 0.06} & 81.5{\tiny{}\textpm 0.23} & 81.9{\tiny{}\textpm 0.05} & 82.8{\tiny{}\textpm 0.19} & 83.1{\tiny{}\textpm 0.17} & 84.0{\tiny{}\textpm 0.42} & 83.9{\tiny{}\textpm 0.18} & 84.5{\tiny{}\textpm 0.07} & 82.2{\tiny{}\textpm 0.28}\tabularnewline
APGD & 16.8{\tiny{}\textpm 0.37} & 8.9{\tiny{}\textpm 0.11} & 8.8{\tiny{}\textpm 0.11} & 23.0{\tiny{}\textpm 2.50} & 46.7{\tiny{}\textpm 3.07} & 30.4{\tiny{}\textpm 3.78} & 22.0{\tiny{}\textpm 5.65} & 15.2{\tiny{}\textpm 0.88} & 13.3{\tiny{}\textpm 0.37} & 10.8{\tiny{}\textpm 1.38} & 8.9{\tiny{}\textpm 0.85} & 8.0{\tiny{}\textpm 0.19} & 7.1{\tiny{}\textpm 0.41} & 23.7{\tiny{}\textpm 3.03}\tabularnewline
APGDT & 17.5{\tiny{}\textpm 0.25} & 10.3{\tiny{}\textpm 0.36} & 11.9{\tiny{}\textpm 0.28} & 27.1{\tiny{}\textpm 3.79} & 54.6{\tiny{}\textpm 1.00} & 40.0{\tiny{}\textpm 3.32} & 31.1{\tiny{}\textpm 6.44} & 22.0{\tiny{}\textpm 1.37} & 18.3{\tiny{}\textpm 1.12} & 14.4{\tiny{}\textpm 1.46} & 12.2{\tiny{}\textpm 0.82} & 10.6{\tiny{}\textpm 0.67} & 9.6{\tiny{}\textpm 0.46} & 34.7{\tiny{}\textpm 2.85}\tabularnewline
BIM & 1.1{\tiny{}\textpm 0.12} & 6.5{\tiny{}\textpm 0.53} & 11.0{\tiny{}\textpm 0.43} & 8.3{\tiny{}\textpm 2.46} & 41.5{\tiny{}\textpm 0.83} & 30.8{\tiny{}\textpm 3.08} & 24.3{\tiny{}\textpm 5.84} & 17.9{\tiny{}\textpm 0.76} & 16.7{\tiny{}\textpm 0.47} & 12.2{\tiny{}\textpm 0.75} & 11.0{\tiny{}\textpm 0.72} & 12.2{\tiny{}\textpm 0.66} & 11.8{\tiny{}\textpm 1.38} & 32.2{\tiny{}\textpm 3.78}\tabularnewline
CW & 55.1{\tiny{}\textpm 0.48} & 67.4{\tiny{}\textpm 0.42} & 71.4{\tiny{}\textpm 0.61} & 62.4{\tiny{}\textpm 1.08} & 74.2{\tiny{}\textpm 2.31} & 77.2{\tiny{}\textpm 0.62} & 77.1{\tiny{}\textpm 1.10} & 75.8{\tiny{}\textpm 0.31} & 75.5{\tiny{}\textpm 0.50} & 74.0{\tiny{}\textpm 0.46} & 73.2{\tiny{}\textpm 0.26} & 73.0{\tiny{}\textpm 0.26} & 72.2{\tiny{}\textpm 0.42} & 77.8{\tiny{}\textpm 0.38}\tabularnewline
EOTPGD & 6.2{\tiny{}\textpm 0.13} & 20.8{\tiny{}\textpm 0.47} & 28.5{\tiny{}\textpm 1.40} & 14.6{\tiny{}\textpm 1.62} & 38.7{\tiny{}\textpm 0.60} & 34.2{\tiny{}\textpm 1.93} & 31.8{\tiny{}\textpm 2.91} & 29.1{\tiny{}\textpm 0.34} & 28.7{\tiny{}\textpm 0.92} & 28.5{\tiny{}\textpm 0.55} & 28.9{\tiny{}\textpm 1.08} & 30.4{\tiny{}\textpm 0.63} & 31.5{\tiny{}\textpm 1.21} & 35.3{\tiny{}\textpm 0.90}\tabularnewline
FFGSM & 6.6{\tiny{}\textpm 0.39} & 21.9{\tiny{}\textpm 0.62} & 26.6{\tiny{}\textpm 1.84} & 11.4{\tiny{}\textpm 0.83} & 32.0{\tiny{}\textpm 0.75} & 27.6{\tiny{}\textpm 1.16} & 26.4{\tiny{}\textpm 1.88} & 24.9{\tiny{}\textpm 0.08} & 26.3{\tiny{}\textpm 1.20} & 28.1{\tiny{}\textpm 0.51} & 29.2{\tiny{}\textpm 0.75} & 32.8{\tiny{}\textpm 0.48} & 34.7{\tiny{}\textpm 1.24} & 32.6{\tiny{}\textpm 1.93}\tabularnewline
FGSM & 18.8{\tiny{}\textpm 0.81} & 40.7{\tiny{}\textpm 0.65} & 45.4{\tiny{}\textpm 1.18} & 31.5{\tiny{}\textpm 2.78} & 59.2{\tiny{}\textpm 2.18} & 55.3{\tiny{}\textpm 1.69} & 52.6{\tiny{}\textpm 3.74} & 49.4{\tiny{}\textpm 0.73} & 48.8{\tiny{}\textpm 1.50} & 48.1{\tiny{}\textpm 0.59} & 48.2{\tiny{}\textpm 0.26} & 50.2{\tiny{}\textpm 0.46} & 50.9{\tiny{}\textpm 0.83} & 56.5{\tiny{}\textpm 1.71}\tabularnewline
GN & 56.0{\tiny{}\textpm 0.42} & 74.8{\tiny{}\textpm 0.33} & 76.5{\tiny{}\textpm 0.40} & 60.1{\tiny{}\textpm 0.35} & 69.7{\tiny{}\textpm 2.34} & 74.4{\tiny{}\textpm 0.28} & 76.4{\tiny{}\textpm 0.34} & 76.7{\tiny{}\textpm 0.02} & 77.6{\tiny{}\textpm 0.39} & 78.6{\tiny{}\textpm 0.09} & 78.9{\tiny{}\textpm 0.43} & 79.4{\tiny{}\textpm 0.22} & 80.2{\tiny{}\textpm 0.22} & 77.7{\tiny{}\textpm 0.21}\tabularnewline
MIFGSM & 0.1{\tiny{}\textpm 0.02} & 0.9{\tiny{}\textpm 0.26} & 1.7{\tiny{}\textpm 0.16} & 1.1{\tiny{}\textpm 0.36} & 16.6{\tiny{}\textpm 0.95} & 8.9{\tiny{}\textpm 2.21} & 5.7{\tiny{}\textpm 2.55} & 3.1{\tiny{}\textpm 0.07} & 3.0{\tiny{}\textpm 0.17} & 2.1{\tiny{}\textpm 0.16} & 1.9{\tiny{}\textpm 0.13} & 2.4{\tiny{}\textpm 0.06} & 2.5{\tiny{}\textpm 0.28} & 9.2{\tiny{}\textpm 2.15}\tabularnewline
PGD & 0.1{\tiny{}\textpm 0.01} & 0.8{\tiny{}\textpm 0.32} & 1.6{\tiny{}\textpm 0.09} & 1.2{\tiny{}\textpm 0.38} & 18.4{\tiny{}\textpm 1.74} & 9.7{\tiny{}\textpm 2.29} & 6.1{\tiny{}\textpm 3.64} & 2.5{\tiny{}\textpm 0.35} & 2.3{\tiny{}\textpm 0.06} & 1.5{\tiny{}\textpm 0.05} & 1.4{\tiny{}\textpm 0.38} & 1.6{\tiny{}\textpm 0.16} & 1.3{\tiny{}\textpm 0.28} & 9.4{\tiny{}\textpm 2.83}\tabularnewline
PGDL2 & 20.5{\tiny{}\textpm 1.19} & 42.7{\tiny{}\textpm 0.75} & 50.3{\tiny{}\textpm 0.86} & 40.3{\tiny{}\textpm 3.49} & 63.8{\tiny{}\textpm 2.37} & 62.6{\tiny{}\textpm 2.13} & 60.7{\tiny{}\textpm 4.03} & 57.9{\tiny{}\textpm 1.16} & 57.8{\tiny{}\textpm 1.34} & 55.5{\tiny{}\textpm 0.43} & 54.8{\tiny{}\textpm 0.26} & 55.2{\tiny{}\textpm 0.24} & 53.2{\tiny{}\textpm 1.29} & 64.8{\tiny{}\textpm 1.60}\tabularnewline
RFGSM & 5.1{\tiny{}\textpm 0.18} & 18.7{\tiny{}\textpm 0.65} & 23.2{\tiny{}\textpm 2.16} & 9.0{\tiny{}\textpm 0.39} & 27.6{\tiny{}\textpm 0.75} & 24.2{\tiny{}\textpm 1.42} & 22.7{\tiny{}\textpm 1.94} & 21.4{\tiny{}\textpm 0.28} & 22.6{\tiny{}\textpm 1.04} & 24.3{\tiny{}\textpm 0.42} & 25.4{\tiny{}\textpm 0.90} & 28.7{\tiny{}\textpm 0.59} & 31.0{\tiny{}\textpm 1.40} & 28.3{\tiny{}\textpm 1.57}\tabularnewline
TPGD & 25.1{\tiny{}\textpm 0.51} & 28.0{\tiny{}\textpm 0.53} & 32.6{\tiny{}\textpm 1.01} & 31.5{\tiny{}\textpm 1.90} & 54.3{\tiny{}\textpm 0.47} & 46.2{\tiny{}\textpm 2.03} & 41.3{\tiny{}\textpm 4.65} & 37.5{\tiny{}\textpm 0.79} & 35.9{\tiny{}\textpm 0.60} & 35.1{\tiny{}\textpm 0.58} & 35.0{\tiny{}\textpm 0.41} & 36.3{\tiny{}\textpm 0.47} & 36.6{\tiny{}\textpm 1.03} & 48.5{\tiny{}\textpm 2.04}\tabularnewline
\hline
\end{tabular}}
\end{small}
\end{center}

\newpage
\par\noindent\refstepcounter{table}\textbf{Table \thetable}\label{tab:adversarial_attack_all_models_tinted}\par
\noindent Comparison of robustness against 12 adversarial attacks and one Gaussian-noise baseline for FDT and DeiT models trained on Tinted ImageNet-100 dataset. The models are labeled with their corresponding fixation budget hyperparameter (subscript) and model size (T, S, B for tiny, small, and base, respectively). The mean and one standard deviation of three runs with different initializations are reported.\par\vspace{0.5em}
\begin{center}
\begin{small}
\setlength{\tabcolsep}{1.7pt}
\resizebox{\textwidth}{!}{%
\begin{tabular}{|l|c|ccccccccc|}
\hline
Attack Type & DeiT-S & FDT-S$_{0.2}$ & FDT-S$_{0.3}$ & FDT-S$_{0.4}$ & FDT-S$_{0.5}$ & FDT-S$_{0.6}$ & FDT-S$_{0.7}$ & FDT-S$_{0.8}$ & FDT-S$_{0.9}$ & FDT-S$_{1.0}$\tabularnewline
\hline
CLEAN & 53.5{\tiny{}\textpm 0.16} & 44.8{\tiny{}\textpm 1.07} & 51.4{\tiny{}\textpm 0.56} & 54.0{\tiny{}\textpm 0.29} & 56.5{\tiny{}\textpm 0.53} & 57.9{\tiny{}\textpm 0.29} & 59.5{\tiny{}\textpm 0.19} & 61.1{\tiny{}\textpm 0.34} & 61.8{\tiny{}\textpm 0.72} & 62.4{\tiny{}\textpm 0.18}\tabularnewline
APGD & 24.9{\tiny{}\textpm 1.27} & 33.0{\tiny{}\textpm 1.43} & 28.5{\tiny{}\textpm 1.45} & 26.1{\tiny{}\textpm 1.74} & 25.1{\tiny{}\textpm 1.10} & 23.0{\tiny{}\textpm 0.77} & 22.0{\tiny{}\textpm 1.15} & 22.0{\tiny{}\textpm 1.09} & 21.5{\tiny{}\textpm 0.54} & 22.5{\tiny{}\textpm 1.39}\tabularnewline
APGDT & 27.5{\tiny{}\textpm 1.08} & 36.6{\tiny{}\textpm 1.45} & 33.4{\tiny{}\textpm 1.67} & 30.3{\tiny{}\textpm 2.11} & 28.7{\tiny{}\textpm 1.38} & 26.1{\tiny{}\textpm 0.93} & 24.5{\tiny{}\textpm 1.57} & 24.4{\tiny{}\textpm 0.73} & 24.3{\tiny{}\textpm 0.72} & 25.2{\tiny{}\textpm 1.33}\tabularnewline
BIM & 6.7{\tiny{}\textpm 0.31} & 21.6{\tiny{}\textpm 2.78} & 16.8{\tiny{}\textpm 1.05} & 12.5{\tiny{}\textpm 1.80} & 9.8{\tiny{}\textpm 0.79} & 7.1{\tiny{}\textpm 0.45} & 5.8{\tiny{}\textpm 0.65} & 6.5{\tiny{}\textpm 0.27} & 6.1{\tiny{}\textpm 0.11} & 7.4{\tiny{}\textpm 0.29}\tabularnewline
CW & 50.6{\tiny{}\textpm 0.48} & 44.1{\tiny{}\textpm 1.02} & 50.0{\tiny{}\textpm 0.63} & 52.2{\tiny{}\textpm 0.76} & 54.1{\tiny{}\textpm 0.83} & 55.2{\tiny{}\textpm 0.26} & 56.2{\tiny{}\textpm 0.67} & 57.5{\tiny{}\textpm 0.23} & 58.6{\tiny{}\textpm 0.62} & 58.9{\tiny{}\textpm 0.43}\tabularnewline
\hline
\end{tabular}}
\end{small}
\end{center}

\newpage

\begin{landscape}
\section{Natural Corruption Results}

\begingroup
\tiny
\setlength{\tabcolsep}{0.4pt}
\begin{longtable}[c]{|l|l|c|ccc|ccccccccccc|}
\caption{Comparison of robustness against natural corruptions at five different severities for FDT and DeiT models trained on ImageNet-100 dataset. Models are evaluated using the ImageNet100-C dataset. The models are labeled with their corresponding fixation budget hyperparameter (subscript) and model size (T, S, B for tiny, small, and base, respectively). The mean and one standard deviation of three runs with different initializations are reported.}
\label{tab:corruption_all}
\\\hline
\multicolumn{2}{|l|}{Corruption} & S  & DeiT-T & DeiT-S & DeiT-B & FDT-T$_{0.5}$ & FDT-S$_{0.2}$ & FDT-S$_{0.3}$ & FDT-S$_{0.4}$ & FDT-S$_{0.5}$ & FDT-S$_{0.6}$ & FDT-S$_{0.7}$ & FDT-S$_{0.8}$ & FDT-S$_{0.9}$ & FDT-S$_{1.0}$ & FDT-B$_{0.5}$\tabularnewline
\hline
\endfirsthead
\caption[]{Comparison of robustness against natural corruptions on ImageNet-100 (continued).}
\\\hline
\multicolumn{2}{|l|}{Corruption} & S  & DeiT-T & DeiT-S & DeiT-B & FDT-T$_{0.5}$ & FDT-S$_{0.2}$ & FDT-S$_{0.3}$ & FDT-S$_{0.4}$ & FDT-S$_{0.5}$ & FDT-S$_{0.6}$ & FDT-S$_{0.7}$ & FDT-S$_{0.8}$ & FDT-S$_{0.9}$ & FDT-S$_{1.0}$ & FDT-B$_{0.5}$\tabularnewline
\hline
\endhead
\multicolumn{2}{|l|}{--} & 0  & 64.4{\tiny{}\textpm 0.48} & 80.9{\tiny{}\textpm 0.11} & 81.5{\tiny{}\textpm 0.25} & 67.1{\tiny{}\textpm 0.08} & 75.4{\tiny{}\textpm 2.55} & 80.0{\tiny{}\textpm 0.06} & 81.5{\tiny{}\textpm 0.23} & 81.9{\tiny{}\textpm 0.05} & 82.8{\tiny{}\textpm 0.19} & 83.1{\tiny{}\textpm 0.17} & 84.0{\tiny{}\textpm 0.42} & 83.9{\tiny{}\textpm 0.18} & 84.5{\tiny{}\textpm 0.07} & 82.2{\tiny{}\textpm 0.28}\tabularnewline
\hline
Blur & \multirow{5}{*}{Defocus Blur} & 1  & 44.6{\tiny{}\textpm 0.05} & 63.5{\tiny{}\textpm 1.04} & 62.8{\tiny{}\textpm 0.85} & 46.6{\tiny{}\textpm 0.28} & 50.2{\tiny{}\textpm 3.75} & 60.5{\tiny{}\textpm 0.65} & 62.5{\tiny{}\textpm 0.59} & 64.9{\tiny{}\textpm 0.22} & 66.0{\tiny{}\textpm 0.12} & 67.0{\tiny{}\textpm 0.64} & 68.1{\tiny{}\textpm 0.50} & 68.5{\tiny{}\textpm 0.65} & 69.6{\tiny{}\textpm 0.51} & 62.9{\tiny{}\textpm 1.00}\tabularnewline
    &  & 2  & 38.1{\tiny{}\textpm 0.40} & 56.7{\tiny{}\textpm 0.59} & 56.4{\tiny{}\textpm 1.03} & 38.8{\tiny{}\textpm 0.48} & 39.1{\tiny{}\textpm 3.27} & 51.8{\tiny{}\textpm 1.04} & 54.3{\tiny{}\textpm 0.31} & 57.1{\tiny{}\textpm 0.82} & 58.8{\tiny{}\textpm 0.38} & 60.1{\tiny{}\textpm 0.65} & 61.7{\tiny{}\textpm 0.97} & 62.6{\tiny{}\textpm 1.17} & 63.5{\tiny{}\textpm 0.58} & 55.3{\tiny{}\textpm 0.79}\tabularnewline
    &  & 3  & 28.4{\tiny{}\textpm 0.33} & 44.0{\tiny{}\textpm 0.03} & 44.7{\tiny{}\textpm 1.00} & 27.4{\tiny{}\textpm 0.25} & 21.0{\tiny{}\textpm 4.10} & 33.7{\tiny{}\textpm 1.63} & 37.2{\tiny{}\textpm 0.44} & 41.8{\tiny{}\textpm 0.50} & 45.2{\tiny{}\textpm 1.26} & 47.2{\tiny{}\textpm 0.73} & 50.5{\tiny{}\textpm 1.01} & 51.1{\tiny{}\textpm 1.36} & 51.7{\tiny{}\textpm 0.68} & 41.0{\tiny{}\textpm 0.50}\tabularnewline
\hline
\end{longtable}
\endgroup

\newpage

\begingroup
\scriptsize
\setlength{\tabcolsep}{1.7pt}
\begin{longtable}[c]{|l|l|c|c|ccccccccc|}
\caption{Comparison of robustness against natural corruptions at five different severities for FDT and DeiT models trained on Tinted-ImageNet-100 dataset. The models are evaluated using the ImageNet100-C dataset. The models are labeled with their corresponding fixation budget hyperparameter (subscript) and model size (T, S, B for tiny, small, and base, respectively). The mean and one standard deviation of three runs with different initializations are reported.}
\label{tab:corruption_small}
\\\hline
\multicolumn{2}{|l|}{Corruption} & S & DeiT-S & FDT-S$_{0.2}$ & FDT-S$_{0.3}$ & FDT-S$_{0.4}$ & FDT-S$_{0.5}$ & FDT-S$_{0.6}$ & FDT-S$_{0.7}$ & FDT-S$_{0.8}$ & FDT-S$_{0.9}$ & FDT-S$_{1.0}$\tabularnewline
\hline
\endfirsthead
\caption[]{Comparison of robustness against natural corruptions on Tinted-ImageNet-100 (continued).}
\\\hline
\multicolumn{2}{|l|}{Corruption} & S & DeiT-S & FDT-S$_{0.2}$ & FDT-S$_{0.3}$ & FDT-S$_{0.4}$ & FDT-S$_{0.5}$ & FDT-S$_{0.6}$ & FDT-S$_{0.7}$ & FDT-S$_{0.8}$ & FDT-S$_{0.9}$ & FDT-S$_{1.0}$\tabularnewline
\hline
\endhead
\multicolumn{2}{|l|}{--} & 0 & 53.5{\tiny{}\textpm 0.16} & 44.8{\tiny{}\textpm 1.07} & 51.4{\tiny{}\textpm 0.56} & 54.0{\tiny{}\textpm 0.29} & 56.5{\tiny{}\textpm 0.53} & 57.9{\tiny{}\textpm 0.29} & 59.5{\tiny{}\textpm 0.19} & 61.1{\tiny{}\textpm 0.34} & 61.8{\tiny{}\textpm 0.72} & 62.4{\tiny{}\textpm 0.18}\tabularnewline
\hline
Blur & \multirow{5}{*}{Defocus Blur} & 1 & 37.7{\tiny{}\textpm 0.67} & 25.7{\tiny{}\textpm 1.12} & 31.7{\tiny{}\textpm 0.73} & 36.1{\tiny{}\textpm 0.77} & 38.9{\tiny{}\textpm 0.69} & 41.4{\tiny{}\textpm 1.73} & 42.7{\tiny{}\textpm 0.15} & 44.6{\tiny{}\textpm 0.39} & 44.8{\tiny{}\textpm 0.71} & 45.9{\tiny{}\textpm 0.38}\tabularnewline
    &  & 2 & 31.1{\tiny{}\textpm 1.01} & 18.9{\tiny{}\textpm 1.24} & 24.9{\tiny{}\textpm 1.04} & 29.4{\tiny{}\textpm 1.35} & 31.6{\tiny{}\textpm 1.02} & 34.2{\tiny{}\textpm 1.85} & 35.9{\tiny{}\textpm 0.33} & 37.7{\tiny{}\textpm 0.50} & 38.0{\tiny{}\textpm 0.47} & 38.9{\tiny{}\textpm 0.39}\tabularnewline
    &  & 3 & 21.2{\tiny{}\textpm 0.80} & 10.4{\tiny{}\textpm 1.56} & 15.7{\tiny{}\textpm 0.81} & 18.7{\tiny{}\textpm 0.26} & 21.0{\tiny{}\textpm 1.37} & 22.4{\tiny{}\textpm 1.75} & 24.4{\tiny{}\textpm 0.91} & 25.9{\tiny{}\textpm 0.35} & 26.0{\tiny{}\textpm 0.19} & 26.6{\tiny{}\textpm 0.75}\tabularnewline
    &  & 4 & 15.4{\tiny{}\textpm 0.70} & 6.6{\tiny{}\textpm 1.36} & 10.6{\tiny{}\textpm 0.57} & 13.0{\tiny{}\textpm 0.26} & 14.4{\tiny{}\textpm 0.85} & 15.4{\tiny{}\textpm 1.20} & 16.7{\tiny{}\textpm 0.63} & 18.1{\tiny{}\textpm 0.49} & 18.1{\tiny{}\textpm 0.28} & 18.6{\tiny{}\textpm 0.70}\tabularnewline
    &  & 5 & 11.7{\tiny{}\textpm 0.52} & 4.6{\tiny{}\textpm 1.00} & 8.2{\tiny{}\textpm 0.31} & 9.5{\tiny{}\textpm 0.15} & 10.2{\tiny{}\textpm 0.49} & 11.1{\tiny{}\textpm 0.90} & 12.0{\tiny{}\textpm 0.42} & 12.9{\tiny{}\textpm 0.24} & 12.9{\tiny{}\textpm 0.24} & 12.9{\tiny{}\textpm 0.53}\tabularnewline
\hline
\end{longtable}
\endgroup
\end{landscape}
